\documentclass[lettersize,journal]{IEEEtran}
\usepackage{amsmath,amsfonts}
\usepackage{algorithmic}
\usepackage{algorithm}
\usepackage{array}
\usepackage[caption=false,font=normalsize,labelfont=sf,textfont=sf]{subfig}
\usepackage{textcomp}
\usepackage{stfloats}
\usepackage{url}
\usepackage{verbatim}
\usepackage{graphicx}
\usepackage{cite}
\usepackage{nicefrac}
\usepackage[table]{xcolor}
\usepackage{multirow}
\usepackage{amssymb}
\usepackage{booktabs}
\usepackage{hyperref}
\usepackage{adjustbox,lipsum}
\usepackage{footnote}
\usepackage{tablefootnote}
\usepackage{scalerel}
\usepackage{enumitem}
\usepackage{makecell}
\usepackage{romannum}
\usepackage{tablefootnote}
\def\etal{\textit{et~al.}}
\def\ie{\textit{i.e.}}
\def\eg{\textit{e.g.}}

\begin{document}

 \title{Mesh Convolution with Continuous Filters \\for
 3D Surface Parsing}

\author{Huan~Lei,
        Naveed~Akhtar,
        Mubarak~Shah,
        and~Ajmal~Mian % <-this % stops a space
\thanks{H.~Lei is with the School of Computing, The Australian National University. N.~Akhtar and A.~Mian are with the Department of Computer Science and Software Engineering, The University of Western Australia, 35 Stirling
Highway, Crawley, Western Australia, 6009. M.~Shah is with the Center for Research in Computer Vision, University of Central Florida, 4328 Scorpius St. Orlando, USA.
E-mail: dr.huanlei@gmail.com, naveed.akhtar@uwa.edu.au, shah@crcv.ucf.edu, 
%\protect\\ \hspace{15mm}
ajmal.mian@uwa.edu.au.}% <-this % stops a space
}



\maketitle

\begin{abstract}
Geometric feature learning {\color{black}for 3D surfaces is critical for many applications in computer graphics and 3D vision.} 
However, deep learning currently lags in hierarchical modeling of  3D surfaces due to the lack of required operations and/or their efficient implementations. In this paper, we propose a series of modular operations for effective geometric feature learning from 3D triangle meshes. These operations include novel mesh convolutions, efficient mesh decimation and associated mesh (un)poolings. 
{\color{black}Our mesh convolutions exploit spherical harmonics as orthonormal bases to create continuous  convolutional filters.} 
The mesh decimation module is GPU-accelerated and able to process batched meshes on-the-fly, while the (un)pooling operations compute features for up/down-sampled meshes. 
We provide open-source implementation of these operations, collectively termed \textit{Picasso}.  
{\color{black}Picasso supports \textit{heterogeneous} mesh batching and processing. Leveraging its modular operations, we further contribute a novel hierarchical neural network for perceptual parsing of 3D surfaces, named PicassoNet++.} It achieves highly competitive performance for shape analysis and scene segmentation on prominent 3D benchmarks. The code, data and trained models are available at \href{https://github.com/EnyaHermite/Picasso}{https://github.com/EnyaHermite/Picasso}.
\end{abstract}


\section{Introduction}\label{sec:intro}

\IEEEPARstart{D}{iscriminative} feature learning for 3D surfaces is fundamentally important for computer graphics and computer vision. 
Although deep learning is able to learn impressive features on images and videos \cite{krizhevsky2017imagenet,he2016deep,liu2016ssd}, {\color{black}its application does not generalize to 
data structures that are not homogeneous grids, such as 3D surfaces. This motivates the research direction of geometric deep learning~\cite{bronstein2017geometric}, which targets irregular data representations, \ie,~graphs, surfaces,  as inputs of neural networks.}

{\color{black}
In digital devices, 3D surfaces are usually represented as discrete polygon meshes, or more commonly triangle meshes. Triangle meshes hold key information about the topology of their smooth counterpart surfaces. We can, therefore, learn features for perceptual parsing of 3D surfaces from such mesh representations. 
However, the variety and heterogeneous nature of atomic components, \ie,~vertices, edges, facets, in meshes makes the ideal feature learning  challenging~\cite{hanocka2019meshcnn,schult2020dualconvmesh,lei2021picasso}.}

{\color{black}
Existing mesh-based 
neural networks mostly learn features for shape analysis \cite{boscaini2016learning,hanocka2019meshcnn,monti2017geometric,ranjan2018generating}. 
Those methods handle small shape meshes as graphs and learn feature representations using graph convolutions. The mesh resolutions in their architectures are either fixed or reduced via inefficient mesh decimation algorithm \cite{garland1999quadric,garland1997surface,rossignac1993multi,zhou2018open3d}.
Whereas non-hierarchical networks and slow resolution reduction are acceptable for small-scale surface analysis like 3D shapes, they become impractical in large-scale surface parsing of 3D scenes. Pioneering works for large-scale surface parsing divide the intact meshes into smaller blocks and learn features with graph convolution~\cite{schult2020dualconvmesh}. 
Currently, applying deep learning to intact scene surfaces is hindered by the absence of amenable mesh convolutions and efficient mesh decimation in the modern libraries \eg,~Pytorch~\cite{paszke2019pytorch}, Tensorflow~\cite{abadi2016tensorflow}.}  

\begin{figure*}[!t]
\centering
\includegraphics[width=0.9\textwidth]{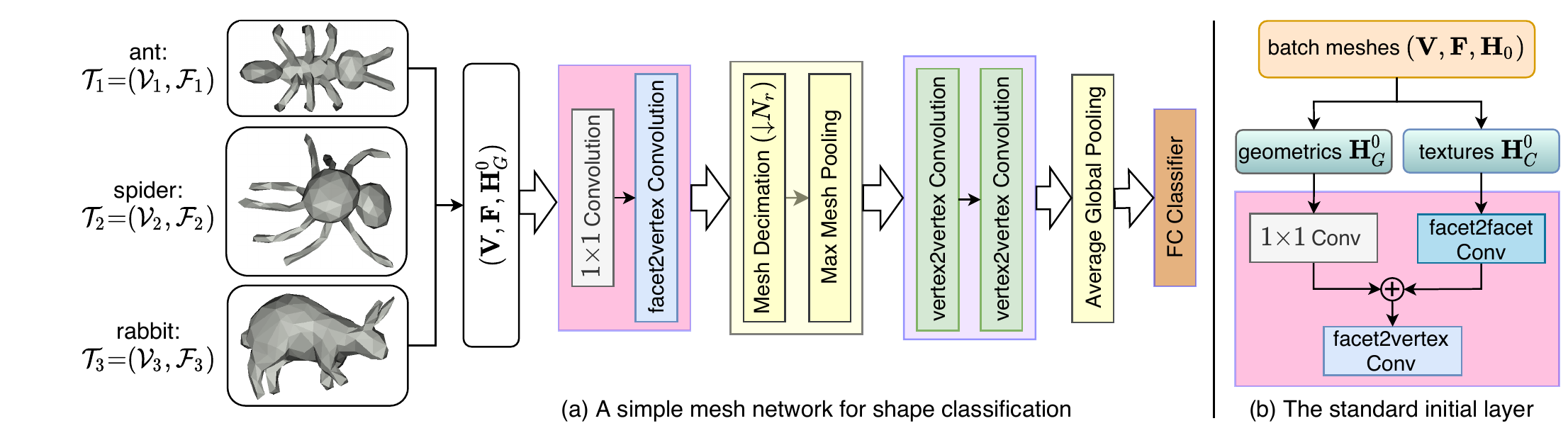}
\vspace{-3mm}
\caption{(a) An example of building a simple hierarchical mesh network for shape classification using the mesh convolutions and poolings in Picasso. The network comprises  two hierarchical layers, and uses batch size 3 in this example. It accepts batch input as a tuple of $({\bf V}, {\bf F}, {\bf H}_G^0)$, where ${\bf V}$ denotes concatenated vertices, ${\bf F}$ are facets, and ${\bf H}_G^0$ denotes geometry features of the shapes. In this illustration, we decimate the input meshes by reducing their number of vertices by $N_r$. (b) Example configuration of the standard initial layer which considers the input features to comprise both geometry features, ${\bf H}_G^0$, and  texture features, ${\bf H}_C^0$. We discuss further details of this figure in the overview of Picasso in  \S~\ref{sec:overview}.}
\label{fig:picasso_demo_usage}
\vspace{-2mm}
\end{figure*}

{\color{black}
With this work, we aim to make deep learning more accessible for 3D surface parsing. To achieve this, we propose novel modular operations that suit triangle meshes such as mesh convolutions with continuous filters, GPU-accelerated mesh decimation, and mesh (un)poolings. 
Specifically, we use spherical harmonics to formulate the convolutional filters as continuous functions of azimuth and elevation angles $(\theta,\phi)$. This 
simplifies the construction of all convolutional kernels in our proposed mesh convolutions, including \textit{facet2vertex}, \textit{vertex2facet}, and \textit{facet2facet} convolution. 
Our formulation parameterizes the angular arguments with face normals for the facet2vertex convolution, while using projected Barycentric coordinates for the vertex2facet and facet2facet convolutions. In addition, we extend the idea of continuous filter modelling to point cloud convolution in ambient spheres by combining spherical harmonics with a function of radius $r$.} Our GPU-accelerated mesh decimation can simplify a batch of meshes on-the-fly for hierarchical feature learning.
It allows control over the decimated mesh resolution using the number of vertices. 
We define the (un)pooling operations based on vertex clusters recorded during the decimation. They are required to generate features for newly-created neurons when the network resolution is altered. 
We implement the presented modular operations in CUDA, and introduce them collectively as \textit{Picasso}. Picasso supports feature learning from heterogeneous
meshes in both PyTorch and Tensorflow.
Figure~\ref{fig:picasso_demo_usage} illustrates an example mesh-based neural network for shape classification
that can be built using the Picasso modules. 

{\color{black}Based on the modular operations, we further contribute a hierarchical mesh-based neural network for semantic analysis of 3D shapes and reconstructed scene surfaces. The proposed network, named PicassoNet++, accepts intact triangle meshes (with or without texture) as inputs.  
PicassoNet++ incorporates a series of significant improvements over the original PicassoNet~\cite{lei2021picasso}, including reduced utilization of point cloud convolutions, increased network depth, and different skip connections.} Our experiments show that dual convolutions, as used in~\cite{lei2020spherical}, are unnecessary for high-resolution meshes as mesh convolutions are effective enough to learn distinctive features for them. This leads to significant computational savings.
To evaluate PicassoNet++, we follow MeshCNN~\cite{hanocka2019meshcnn} and use the SHREC \cite{lian2011shape}, CUBE \cite{hanocka2019meshcnn}, COSEG \cite{wang2012active}, Human \cite{maron2017convolutional}, and FAUST \cite{bogo2014faust} datasets for shape analysis. We also use the large-scale ShapeNetCore dataset~\cite{chang2015shapenet} for 3D shape classification, and the S3DIS \cite{armeni20163d} and ScanNet \cite{dai2017scannet} datasets for large-scale surface segmentation. Our approach achieves highly competitive results in all cases.

This article is a significant extension of our preliminary work presented in IEEE CVPR 2021~\cite{lei2021picasso}. 
The following outlines our main contributions  beyond the conference work.
\begin{itemize}
\item {\color{black}\textbf{Filter modelling with spherical harmonics:} We simplify the construction of mesh convolutional kernels by using spherical harmonics. This allows the discrete filters to be represented as continuous real-valued functions, rather than discrete values in $\mathbb{R}$. Since face normals are distributed on the surface of a unit sphere, we model filters for \textit{facet2vertex} convolutions as continuous functions of their azimuth and elevation angles $(\theta, \phi)$. This is superior to the fuzzy modeling of using mixture models \cite{lei2021picasso} because spherical harmonics are fixed functions and do not require additional training parameters. We also apply this idea to \textit{vertex2facet} and \textit{facet2facet} convolutions by computing the angular arguments $(\theta, \phi)$ from projected Barycentric coordinates. Additionally, we combine the spherical harmonics with a radial function of radius $r$ to construct continuous filters for point cloud convolution in ambient spheres.

\vspace{0.5mm}
\item \textbf{PicassoNet++:} We present a novel mesh based neural network that can be applied for general 3D surface parsing from their triangle mesh representations. By establishing the passive role of point cloud convolutions in high-resolution mesh feature learning, we address it in PicassoNet++ to gain a significant computational advantage over \cite{lei2021picasso} while maintaining the performance. We further improve the network architecture through better design choices of skip connections and sub-network blocks. PicassoNet++ accepts intact meshes as inputs, and is applicable to textured meshes.
We incorporate all required functionalities in our network, including reconfiguration of the initial convolutional layer - see Fig.~\ref{fig:picasso_demo_usage}.} 
\vspace{0.5mm}
\item \textbf{Extensive evaluation:} We evaluate PicassoNet++ on various 3D benchmarks for surface parsing, including small-scale shape analysis and large-scale scene segmentation. It achieves highly competitive performance on all datasets. We also provide ablation studies for determining the neural network architecture and the truncated degree of spherical harmonics. We release the latest Picasso and PicassoNet++ at \href{https://github.com/EnyaHermite/Picasso}{https://github.com/EnyaHermite/Picasso} for the broader research community.
\item \textbf{Pytorch extension:} Originally in \cite{lei2021picasso}, Picasso was only supported in Tensorflow. However, with this work, we also make it available in PyTorch due to the growing popularity of this community. We note that the Picasso released with this article incorporates not only the newly introduced modules for heterogeneous mesh processing, but also compatible modular operations for heterogeneous point cloud processing. We include the point cloud modules by adapting our previous contributions~\cite{lei2020spherical,lei2020seggcn} to heterogeneous applications. Altogether, Picasso enables convenient building of neural networks for heterogeneous mesh processing in PyTorch and Tensorflow.
\end{itemize}

\section{Related Work}\label{sec:references}
{\color{black}
We review convolutional methods for 3D triangle meshes, as well as convolutions over 3D point clouds as they can be applied to the vertices of a triangle mesh. Finally, we discuss the existing algorithms for mesh decimation.

\vspace{-1mm}
\subsection{Convolution on 3D Meshes}\label{subsec:convolution_mesh}
Multiple approaches exist that employ convolution on meshes to learn features for small-scale shape analysis. 
The convolutions are generally performed  on local planar patches defined in the hand-crafted coordinate systems \cite{boscaini2016learning,masci2015geodesic,monti2017geometric}. These methods either establish the coordinate system using  geodesic level sets \cite{masci2015geodesic}
or surface normals and principle curvatures \cite{boscaini2016learning,monti2017geometric}.
For improved correspondence matching, Verma \etal \cite{verma2018feastnet} replaced the previous hand-crafted local patches with a learnable
mapping between graph neighborhoods and filter weights.
To reconstruct human facial expressions, Ranjan~\etal~\cite{ranjan2018generating} exploited the spectral graph convolutions \cite{defferrard2016convolutional} with hierarchical mesh-based autoencoders. Gao~\etal~\cite{gao2022robust} proposed to learn mesh representations with adaptive attention matrices and anisotropic convolutions.

Whereas other methods focus on learning vertex-wise features,
MeshCNN~\cite{hanocka2019meshcnn} introduces convolutional operation that learns edge-wise features for semantic labelling on a mesh. The recent PD-MeshNet~\cite{milano2020primal} further extracts facet-wise representations by defining convolution on the primal-dual graphs of an input mesh. It reduces network resolution using the graph edge contraction method provided by Pytorch Geometric \cite{fey2019fast}. HodgeNet~\cite{smirnov2021hodgenet} proposes to learn the mesh geometry from the spectral domain, which involves computing eigenvalues/eigenvectors and can be time-consuming. SubdivNet~\cite{hu2022subdivision} uses loop subdivisions to learn features from meshes, whereas the input mesh has to be remeshed~\cite{liu2020neural} for fine connectivity.

Currently, only a small number of mesh-based convolutional networks exist for large-scale scene parsing in the real world. TextureNet \cite{huang2019texturenet} parameterizes the room surface into local planar patches in the 4-RoSy field such that standard CNNs \cite{krizhevsky2012imagenet} can be applied to extract high-resolution texture information from mesh facets. Schult~\etal   \cite{schult2020dualconvmesh} applied the spatial graph convolutions of dynamic filters \cite{li2018pointcnn,simonovsky2017dynamic,wang2018dynamic,wu2019pointconv} to the union of  neighborhoods in both geodesic and Euclidean domains for vertex-wise feature learning. VMNet \cite{hu2021vmnet} combines the SparseConvNet \cite{tang2020searching} with graph convolutional networks to learn merged features from point clouds and meshes. 
Generally, previous methods explore mesh as an edge-based graph and define the graph convolutions based on its geodesic connections \cite{milano2020primal,verma2018feastnet,schult2020dualconvmesh,ranjan2018generating}. We instead propose convolutions on the mesh structure itself, following its elementary geometric components, \ie,~vertices and facets. To promote this more natural perspective, we also contribute computation and memory optimized CUDA implementations 
for forward and backward propagations of all the mesh convolutions we present in this work. 

\vspace{-1mm}
\subsection{Convolution on 3D Point Clouds}\label{subsec:convolution_pointcloud}
Applying voxel-grid kernels to dense volumetric representations is the most straightforward solution of transferring CNNs from images to point clouds \cite{wu20153d,maturana2015voxnet,zeng20163dmatch}.
However, the practical potential of these methods is limited by their cubically growing requirements on memory and computational resources. 
Different strategies have been introduced to incorporate sparsity into the dense volumetric CNNs \cite{EngelckeICRA2017,graham20183d,choy20194d,riegler2017octnet}, among which SparseConvNets \cite{graham20183d,choy20194d} are currently the best performing architectures.  
Several approaches also explore similar regular-grid kernels for transformed input representations of point clouds, such as TangentConv \cite{tatarchenko2018tangent}, SplatNet \cite{su2018splatnet},  UnPNet~\cite{li2021rethinking}.
Since PointNet \cite{qi2017pointnet}, the permutation invariant networks learn features from point clouds using multilayer perceptrons followed by max pooling~\cite{klokov2017escape,qi2017pointnet,qi2017pointnetplusplus,wu2019pointconv} and the spatial coordinates of points are used as input features.

Graph-based neural networks allow the
convolutions to be conducted in either spectral or spatial domain. However, applying the spectral convolutions to point cloud processing is complicated because they demand the graph Laplacians of different input samples to be pre-aligned \cite{yi2017syncspeccnn}. 
As a pioneering work in the spatial domain, ECC \cite{simonovsky2017dynamic} exploits dynamical filters \cite{de2016dynamic} to 
parameterize the graph convolutional parameters for point cloud analysis. Subsequent works also explored more effective kernel and filter parameterizations \cite{li2018pointcnn,wang2019attention,wu2019pointconv,du2022novel,huang2022dual}.
The discrete kernels  \cite{lei2019octree,lei2020seggcn,lei2020spherical,thomas2019kpconv} are efficient alternatives to those dynamic kernels as they define the filter parameters directly, avoiding the necessity of indirect filter generation within the network. 
The spherical kernels \cite{lei2020seggcn,lei2020spherical} that separate depth-wise and point-wise computations are memory and runtime advantageous, while KPConv \cite{thomas2019kpconv} is reported to be more competitive than SparseConvNets. 
We refer interested readers to surveys~\cite{wu2020comprehensive, li2020deep} for progress in deep learning for graph neural networks and point clouds. 
Recently, researchers have also started to adapt transformers~\cite{vaswani2017attention} to point cloud processing~\cite{zhao2021point, liu2022uninet, guo2021pct, wu2022point}.}

\vspace{-1mm}
\subsection{Mesh Decimation}\label{subec:decimation_mesh}
Hierarchical neural networks induce  multi-scale feature extraction by allowing convolutions to be applied on increasing receptive fields of the input data. Although farthest point sampling (FPS) is widely used to construct hierarchical architectures for point clouds \cite{lei2020spherical,qi2017pointnetplusplus,wu2019pointconv}, it is inapplicable to mesh processing because of its inability of tracking vertex connections. 
Fortunately, the graphics research community has contributed effective  methods for mesh simplification, such as Vertex Clustering (VC) \cite{rossignac1993multi} and Quadric Error Metrics (QEM)~\cite{garland1999quadric,garland1997surface}. 
The two methods are suitable choices for mesh-based neural networks \cite{ranjan2018generating,schult2020dualconvmesh,hu2021vmnet} to establish hierarchical architectures. Compared to VC, the QEM method is good at reducing mesh resolution
while retaining most of its geometric information, leading to  superior performance \cite{schult2020dualconvmesh}. In specific, QEM simplifies a mesh via iterative contractions of  vertex pairs, whereas the optimal vertex pair for contraction has to be determined after each iteration. The popular geometric processing library - Open3D  \cite{zhou2018open3d}, offers simplification functions for both VC and QEM. 
However, the  CPU-based implementation is inefficient and also not amenable to operations required for deep learning, \eg,~batch processing. {\color{black}Recently, researchers have also started to explore the possibility of achieving mesh simplification using neural networks~\cite{potamias2022neural}.}

Though superior to VC in performance, the iterative progressive strategy of QEM makes it  impossible to be deployed on GPUs as parallel processes. In this work, we introduce a fast mesh decimation technique based on the QEM algorithm \cite{garland1997surface}. Compatible to deep learning, our method can process a batch of heterogeneous meshes on-the-fly. In contrast to \cite{garland1997surface}, it sorts all the vertex pairs only once according to their quadric errors, and groups the vertices to be contracted into \textit{disjoint} clusters. Except for the grouping process, all other computations in our method get accelerated via parallel GPU computing. 

\begin{figure*}[!t]
\centering
\includegraphics[width=0.98 \textwidth]{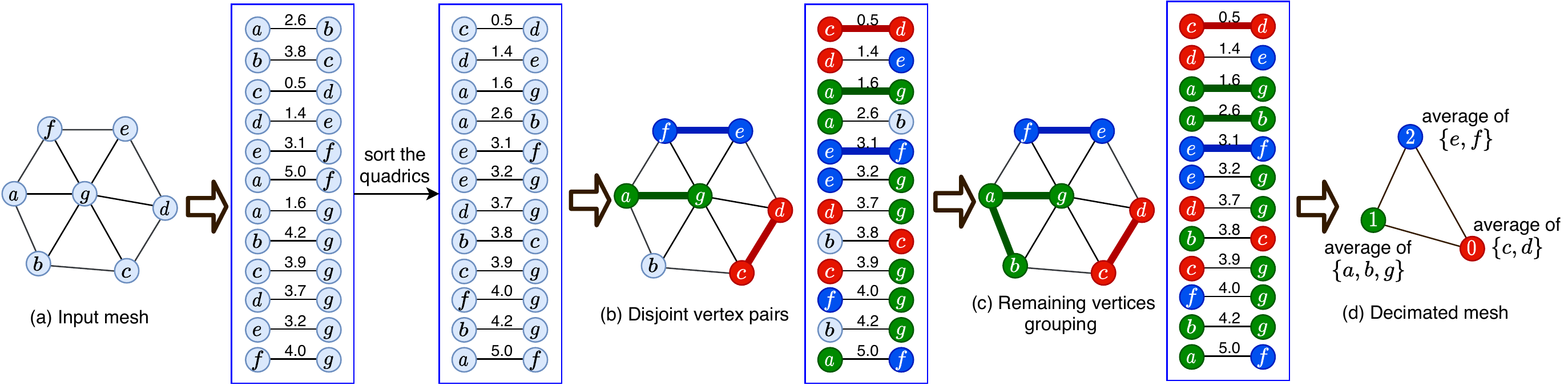}\\
\vspace{-3mm}
\caption{Illustration of the vertex clustering process. (a) An input mesh with twelve edges (vertex pairs). We sort the vertex pairs in ascending order according to their quadric errors. (b) Then, we initialize the clusters as $\{c,d\},\{a,g\},\{e,f\}$ using the disjoint vertex pairs $(c,d),(a,g),(e,f)$ - shown by red, green and blue. (c) We group the remaining 
vertex $b$ to the vertex cluster $\{a,g\}$ because $(a,b)$ holds the smallest quadric error among all pairs containing $b$, \ie $(a,b),(b,g),(b,c)$. Finally, the vertex clusters become $\{c,d\},\{a,b,g\},\{e,f\}$. (d) We construct the decimated mesh by applying vertex contraction to each cluster. 
The target position of contraction is computed as the average location of all vertices in the cluster.} 
\label{fig:vertex_cluster}
\vspace{-2mm}
\end{figure*}

\section{GPU-Accelerated Mesh Decimation}
To explore hierarchical neural networks on 3D meshes, there is a need of  efficient mesh decimation technique that suits deep learning for on-the-fly network reduction. The QEM algorithm \cite{garland1997surface} is effective at simplifying meshes while retaining the decimation quality. However, it applies contractions to each vertex pair iteratively with a global optimal quadric error. 
The implicit dependencies between the iterative contractions make this method unsuitable for parallel acceleration. 
Hence, we propose an enhancement of QEM to enable parallel computing with GPUs. 
In our method, we do not allow inter-dependent iterative contractions. 
Instead, we group the vertices into multiple disjoint clusters under a reasonable compromise on the 
quadric error cost. 
We control the clustering process using expected number of vertices in the decimated mesh rather than the number of edges or facets. 
Due to the disjointness of vertex clusters, their contractions are independent of each other and can be executed in parallel. 
We provide a toy example in Fig.~\ref{fig:vertex_cluster} to illustrate our procedure of vertex clustering.

In our method, we establish the vertex pairs for candidate contraction using the end-vertices of mesh geodesic edges only. 
To prioritize the vertex pairs that contribute to lower quadric errors, they are arranged in ascending order.  
Each vertex cluster is then initialized as a disjoint vertex pair in the ascending order of the candidates.
We summarize our mesh simplification procedure as Algorithm~\ref{alg:Mesh_decimate_ours} that reduces the number of mesh vertices to nearly a half per-iteration.
To handle mesh decimation of arbitrary number of vertices, we allow the core algorithm to be iterated for flexible (${\geqslant}1$) times.
In Algorithm~\ref{alg:Mesh_decimate_ours}, we present the decimation method for a single mesh as the input for clarity. Our decimation function implementation processes `mini-batches' of multiple meshes. 
We execute the vertex clustering (\textit{lines 5--16}) on CPU while all the other operations that require heavy computations are performed on GPU. The clustering process has a time complexity of  $\mathcal{O}(|\mathcal{E}|)$,  where $|\mathcal{E}|$ is the number of edges of the input mesh.
The routine penalties, and consistency checks in mesh decimation are excluded in our method to favor runtime efficiency. We compare the runtime of QEM and the proposed decimation algorithm in Fig.~B.1 of the supplementary, where our method is much faster.

\begin{algorithm}[t]
\caption{The GPU-accelerated mesh simplification}
\label{alg:Mesh_decimate_ours}
\begin{algorithmic}[1]
\renewcommand{\algorithmicrequire}{\textbf{Input:}}
\renewcommand{\algorithmicensure}{\textbf{Output:}}
\REQUIRE mesh $\mathcal{T}^i{=}(\mathcal{V}^i,\mathcal{F}^i)$;
number of vertices to remove $N_r$. \hspace{-4mm}
\ENSURE decimated mesh ${\mathcal{T}^o}{=}(\mathcal{V}^o,\mathcal{F}^o)$.
\vspace{1mm}
\STATE  establish a vertex pair $(v_i$, $v_j)$ for each edge. 
\STATE compute the quadric cost of contracting each pair.
\STATE sort all pairs ascendingly based on the quadrics.
\STATE set $n_r=0$, and $p(v_i)=\FALSE, \forall~v_i \in \mathcal{V}^i$. 
\FOR{each pair ($v_i$, $v_j$)}
\IF{$p(v_i)=\FALSE$, $p(v_j)=\FALSE$, \AND $n_r<N_r$}
\STATE (a) initialize  $\{v_i, v_j\}$ as a new cluster.
\STATE (b) set $n_r=n_r+1$, $p(v_i)=\TRUE$, $p(v_j)=\TRUE$.
\ENDIF
\ENDFOR
\FOR{each pair ($v_i$, $v_j$)}
\IF{$p(v_i)=\FALSE$ \OR $p(v_j)=\FALSE$, \AND $n_r<N_r$}
\STATE (a) place $v_i$, $v_j$ to the same cluster.
\STATE (b) set $p(v_i)=\TRUE$, $p(v_j)=\TRUE$.
\ENDIF
\ENDFOR
\FOR{each cluster $\{v_i, v_j, \dots\}$}
\STATE (a) compute the average position $\bar{v}$ of the cluster.
\STATE (b) contract the cluster to $\bar{v}$.
\ENDFOR
\STATE return
\end{algorithmic}
\footnotetext{We compute $\bar{v}$ as the average position of all vertices in a cluster.}
\end{algorithm}
In our implementation, 
we also record the vertex clustering information with a parameter \textit{VCluster}, and the vertex mapping between input and output meshes with a parameter \textit{IOmap}. They are both vectors of the same sizes as the number of vertices $|\mathcal{V}^i|$ in the input mesh $\mathcal{T}^i=(\mathcal{V}^i,\mathcal{F}^i)$.
Our decimation function yields those two parameters along with the decimated mesh as they are required in the computations of (un)poolings.

\vspace{1mm}
\noindent \textbf{(Un)poolings:} The clustering and mapping information  encoded in vectors \textit{VCluster} and \textit{IOmap} largely facilitate the (un)pooling computations.  
Considering each cluster as a local region or neighborhood, common pooling operations such as `sum'/`average'/`max'/`median'/`weighted' can be directly defined. We provide max($\cdot$), and average($\cdot$) poolings to down-sample the features. For unpooling, all vertices in a cluster replicate features of a representative vertex that the cluster is contracted to in the decimated mesh. 
Consider the input and output meshes in Fig.~\ref{fig:vertex_cluster} as an  example. We compute the feature of vertex `$1$' in the decimated mesh as $h^1=\max(h^a,h^b,h^g)$ under max pooling, while create the features of $\{a,b,g\}$ as $h^a=h^b=h^g=h^1$ in unpooling.
In addition to the mesh decimation and (un)poolings, we also introduce convolutional operations that are more compatible to feature learning on triangular meshes than the previous graph convolutions \cite{schult2020dualconvmesh}.

{\color{black}
\section{Mesh Convolutions}
\vspace{2mm}
We represent a triangle mesh as $\mathcal{T}=(\mathcal{V},\mathcal{F})$, where $\mathcal{V}$ and $\mathcal{F}$ denote the set of vertices and facets respectively. 
Let the spatial coordinates of each vertex $v$ be ${\bf x}$, while the area and normal of a facet $f$ be $A$ and ${\bf n}$.
In the case of a textured mesh, we denote the texture size of each facet as $K \times 3$, where $K$ represents the texture resolution and 3 indicates the dimension of colors. Note that, $K$ varies across different facets.
\begin{figure*}[!t]
    \centering
\includegraphics[width=0.95\textwidth]{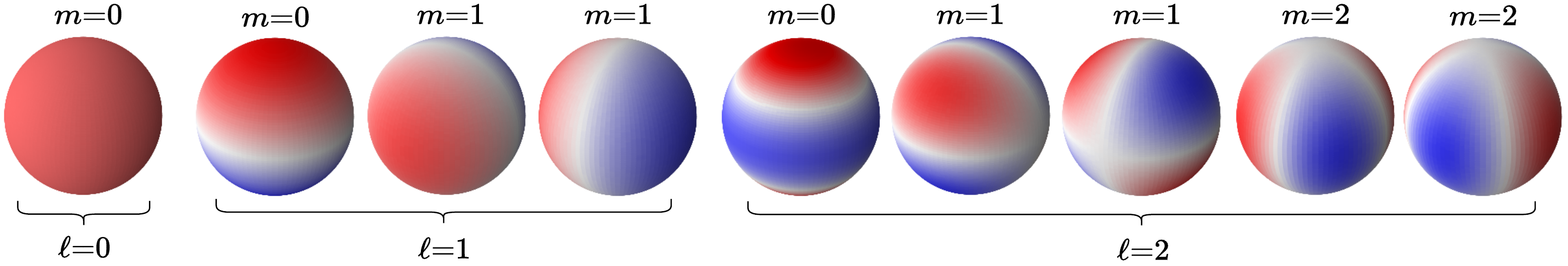}
    \vspace{-3mm}
    \caption{\color{black}Visualization of spherical harmonics of degree $L=2$. We show both the real and imaginary parts for each spherical harmonic $Y_l^m$ with order $m>0$. We approximate each real-valued filter function ${\mathbf F}(\theta,\phi)$ with linear combinations of truncated spherical harmonics. This formulation is applied to parameterize the filters in all of our proposed mesh
convolutions, \ie,  the facet2vertex, vertex2vertex, and
facet2facet convolutions.} 
    \label{fig:sphere_harmonics}
    \vspace{-2mm}
\end{figure*}

Let $\{\mathbf{F}_t\}_{t=1}^T$ be the discrete filters in a convolutional kernel, where $T$ is the kernel size. In deep learning, to compute the feature of a given item $j$ using convolutions, we consider its receptive field $\mathcal{N}(j)$ and apply the following 
\begin{align}
&g_j =\frac{1}{\mathcal{N}(j)}\sum_{i\in\mathcal{N}(j)}\langle\mathbf{F}_{t_i},\mathbf{h}_i\rangle.
\label{eq:basic_Conv}
\end{align}
Here, $t_i$ refers to the filter indexed by its neighboring item $i$ in the receptive field, and $\mathbf{h}_i$ refers to the input features of item $i$. Let the dimension of input features  be  $C$, it then follows that $g_j\in\mathbb{R}$, and ${\mathbf F}_i, {\mathbf h}_i\in\mathbb{R}^C$. 

\subsection{Filter Modelling with Spherical Harmonics}\label{subsec:continuous_filter_ylm}
Discrete kernels~\cite{lei2021picasso, thomas2019kpconv, choy20194d, lei2020spherical} require brute-force partitioning of the subspace defined by local receptive fields. To avoid this, we propose formulating the discrete filters $\{\mathbf{F}_t\}_{t=1}^T$ as a \textit{single continuous} function $\mathbf{F}$, and we parameterize it using the surface normals of 3D meshes. We note that $\mathbf{F}{\in}\mathbb{R}^C$ is a vector function. Inspired by the characteristics of basis functions, we represent the continuous filter as linear combinations of such functions. Since normals are distributed on the surface of a unit sphere, we choose the spherical harmonics $\{Y_l^m\}$ as the basis functions. Thus, the filter function $\mathbf{F}$ is expressed as 
\begin{align}\label{eq:spharm_filter}
\mathbf{F}(\theta,\phi) &= \sum_{\ell=0}^{\infty}\sum_{m=-\ell}^{\ell}\boldsymbol{\alpha}_{\ell m}Y_{\ell}^m(\theta,\phi),
\end{align}
where $\{\boldsymbol{\alpha}_{\ell m}\in\mathbb{C}^C\}$ are the linear coefficients, and 
\begin{align}\label{eq:ylm}
Y_{\ell}^m(\theta,\phi) &= \sqrt{\frac{2\ell+1}{4\pi}\frac{(\ell-m)!}{(\ell+m)!}}P_{\ell}^m(\cos{\theta})e^{im\phi},
\end{align}
for $m\geq0$. The integers $l$ and $m$ indicate the degree and order of a spherical harmonic, respectively. $P_l^m(\cdot)$ are associated Legendre polynomials without the Condon–Shortley phase.    
We denote the azimuthal angle as $\phi \in [0,2\pi)$ and the polar angle as $\theta \in [0,\pi]$, following the common conventions. In practice, we use truncated spherical harmonics with a finite degree $L$ to approximate the continuous filter functions.

Equation (\ref{eq:ylm}) implies that the spherical harmonics satisfy the property $Y_l^{-m}(\theta,\phi){=}(-1)^mY_l^m(\theta,\phi)^*$. Since the filter $\mathbf{F}(\theta,\phi)$ is a \textit{real-valued} function, we can combine each pair of $Y_l^{-m}$ and $Y_l^m$, and represent $\mathbf{F}(\theta,\phi)$ using truncated spherical harmonics as follows
\begin{equation}\label{eq:real_spharm_filter}
\begin{split}
\mathbf{F}(\theta,\phi)= \sum_{\ell=0}^{L}\Big( &\sum_{m=1}^{\ell}\mathbf{a}_{\ell m}Y_{\ell}^m(\theta,0)\cos(m\phi)+\\
&\sum_{m=1}^{\ell}\mathbf{b}_{\ell m}Y_{\ell}^m(\theta,0)\sin(m\phi)+\\
&\mathbf{a}_{\ell 0}Y_{\ell}^0(\theta,\phi)\Big).
\end{split}
\end{equation}
Figure~\ref{fig:sphere_harmonics} shows an example of such a spherical harmonic basis with degree $L{=}2$. In this representation, the learnable parameters in our convolutional kernel are the linear coefficients $\{\mathbf{a}_{\ell 0}, \mathbf{a}_{\ell m}, \mathbf{b}_{\ell m}{\in}\mathbb{R}^C|m{\in}[\ell], \ell{\in}[L]\}$, where $[\ell]{=}\{1,2,\dots,\ell\}$ and similarly $[L]$. We redefine the size of our convolutional kernel $T$ as the number of such coefficients, \ie, the number of basis functions in Eq.~(\ref{eq:real_spharm_filter}). This results in the kernel size being $T=(L+1)^2$.
We apply the filter modelling in Eq.~(\ref{eq:real_spharm_filter}) to all of our proposed mesh convolutions, including the \textit{facet2vertex}, \textit{vertex2vertex}, and \textit{facet2facet} convolutions. 

\subsection{Angular Arguments in Different Convolutions}
\noindent\textbf{Facet2vertex convolution.}  We compute features of each vertex by aggregating context information from adjacent facets, rather than neighboring vertices. This avoids transforming a mesh into a graph for context propagation.
The facet normal is directional data residing on the surface of a unit sphere. We, therefore, compute the angular arguments $(\theta,\phi)$ of its filter functions based on facet normals. 

Following previous works \eg~\cite{chollet2017xception,lei2020spherical}, we define the facet2vertex convolution in a depth-wise separable manner to save computations. Let the learnable coefficients in the kernel be $\{a_{\ell 0}, a_{\ell m}, b_{\ell m}{\in}\mathbb{R}|m{\in}[\ell], \ell{\in}[L]\}$, the adjacent facets of vertex $v$ be $\mathcal{N}(v)$, and the associated features of those facets be $\{h_f|{f\in\mathcal{N}(v)}\}$. The feature of vertex $v$ is computed as 
\begin{align}
g_v =\frac{1}{\mathcal{N}(v)}\sum_{f\in\mathcal{N}(v)}F(\theta_f, \phi_f){h_f}.
\label{eq:F2V_Conv}
\end{align}
We use ReLU \cite{nair2010rectified} as the activation function. Considering
our filter modelling based on normals, the facet2vertex convolution is scale and translation invariant but not rotation invariant.

\vspace{1mm}
\noindent\textbf{Vertex2facet convolution.} We aggregate features of each facet from its vertices. The vertex2facet convolution also exploits depth-wise separable strategy. Therefore, its definition of learnable parameters is the same as that for the facet2vertex convolution. We compute angular arguments of its filters from Barycentric coordinates of each  vertex of a facet, which are $(1,0,0)$ for vertex $v_1$, $(0,1,0)$ for vertex $v_2$, and $(0,0,1)$ for vertex $v_3$. 
They correspond to the angular values of  $(\frac{\pi}{2},0)$, $(\frac{\pi}{2},\frac{\pi}{2})$, and $(0,0)$ on the sphere, respectively.

Let $\{h_1,h_2,h_3\}$ be the features of vertices $\{v_1,v_2,v_3\}$. We compute the feature of facet $f$ as
\begin{align}
g_f=F(\frac{\pi}{2},0)h_1+
F(\frac{\pi}{2},\frac{\pi}{2}) h_2+F(0,0)h_3.
\label{eq:V2F_Conv}
\end{align}
The Barycentric interpolation in \cite{lei2021picasso} is no longer retained in the vertex2facet convolution 
as it only makes a minor contribution to  feature extraction but requires additional computations.
To propagate local information from vertices to vertices \cite{qi2017pointnet,wu2019pointconv,lei2020spherical}, we induce a vertex2vertex convolution by combining the vertex2facet and facet2vertex convolutions.  
Figure~\ref{fig:v2v_conv} illustrates the notion of facet2vertex, vertex2facet,   vertex2vertex,
and facet2facet convolutions. 

\vspace{1mm}
\noindent\textbf{Facet2facet convolution.}
When the input mesh is textured, we learn the texture features of each facet based on the colors of all the internal points on the facet. 
Let $\{{\bf h}_k\in\mathbb{R}^3\}$ be the input colors of all points on a facet, and the associated Barycentric coordinates of each point be $\{{\boldsymbol\xi}_k=[\xi_{k1},\xi_{k2},\xi_{k3}]^\intercal|\xi_{k1}+\xi_{k2}+\xi_{k3}=1,~\xi_{k1},\xi_{k2},\xi_{k3}\geqslant0\}$. 
A facet of texture resolution $K$ leads to $|\{{\bf h}_k\}|=|\{{\boldsymbol\xi}_k\}|=K$. 

We do not exploit 
depth-wise separable strategy in the facet2facet convolution since there are only three (color) channels. 
Therefore, its learnable parameters are defined as $\{\mathbf{a}_{\ell 0}, \mathbf{a}_{\ell m}, \mathbf{b}_{\ell m}{\in}\mathbb{R}^3|m{\in}[\ell], \ell{\in}[L]\}$. 
We calculate the angular arguments for its filters by projecting $\{{\boldsymbol\xi}_k\}$ on a simplex to a unit sphere. 
We compute the feature of a facet $f$ whose texture resolution is $K$ as
\begin{align}
g_f&=\frac{1}{K}\sum_k \langle{\mathbf{F}(\theta_k,\phi_k),{\bf h}_k}\rangle.
\label{eq:F2F_Conv}
\end{align}

The facet2facet convolution is only required at the first convolution layer for extracting texture information from the raw mesh.
In the experiments, we utilize Barycentric interpolation~\cite{coxeter1961introduction} to prepare the mesh textures. 
The texture resolution $K$ of a facet ${\bf f}$ is determined by its area $A$, \ie,~ 
\begin{equation}\label{equ:interpolate_num}
K=\frac{(\gamma+1)(\gamma+2)}{2},~\text{where}~ 
\gamma=\left\lfloor \frac{\alpha(A-A_{\min})}{A_{\max}-A_{\min}}  \right\rfloor  + \beta.\\  
\end{equation}
Here, $A_{\min}, A_{\max}$ are the minimum and maximum facet areas of the mesh, whereas  $\alpha,\beta\in\mathbb{Z}_{\scaleto{\geqslant0\mathstrut}{5pt}}$ are  hyper-parameters. 

\subsection{Extension to Point Cloud Convolution}\label{subsec:pcloud_conv}
Point cloud convolution aggregates context information of a point from its nearest neighbors, which are usually constructed using range search \cite{preparata2012computational}. This results in its receptive field being within an ambient sphere rather than on a sphere surface. The filters in its convolutional kernel should be functions of three arguments $(\theta,\phi,r)$, where $r$ indicates the radial variable. For simplicity, we do not use orthonormal basis in 3D space to model the filter $F(\theta,\phi,r)$. Instead, we extend the formulation of $F(\theta,\phi)$ in Eq. (\ref{eq:real_spharm_filter}) by adding radial controls. It is noted  that the Wigner D-functions \cite{cohenspherical}, which are defined for SO(3) of Euler angles, are not applicable in our case.

Similar to the facet2vertex and vertex2facet convolutions, we exploit depth-wise separable strategy in the point cloud convolution. Finally, the filter $F(\theta,\phi,r)$ is defined as 
\begin{align}
F(\theta,\phi,r) = F(\theta,\phi)Z(r) + c_{0}\big(1-Z(r)\big),
\end{align}
where $Z(r)=\frac{r}{\rho},~r\in[0,\rho]$ and $\rho$ is the radius of a sphere. This modeling introduces only an additional parameter $c_0$ to the learnable parameters and significantly simplifies the computation. It also nicely correlates $F(\theta,\phi)$ with $F(\theta,\phi,r)$, where $F(\theta,\phi)\equiv F(\theta,\phi,\rho)$. The normalized $Z(r)$ makes it easier to use the filter across different scales and spatial locations in a point cloud.}

\begin{figure}[!t]
    \centering
\includegraphics[width=0.49\textwidth]{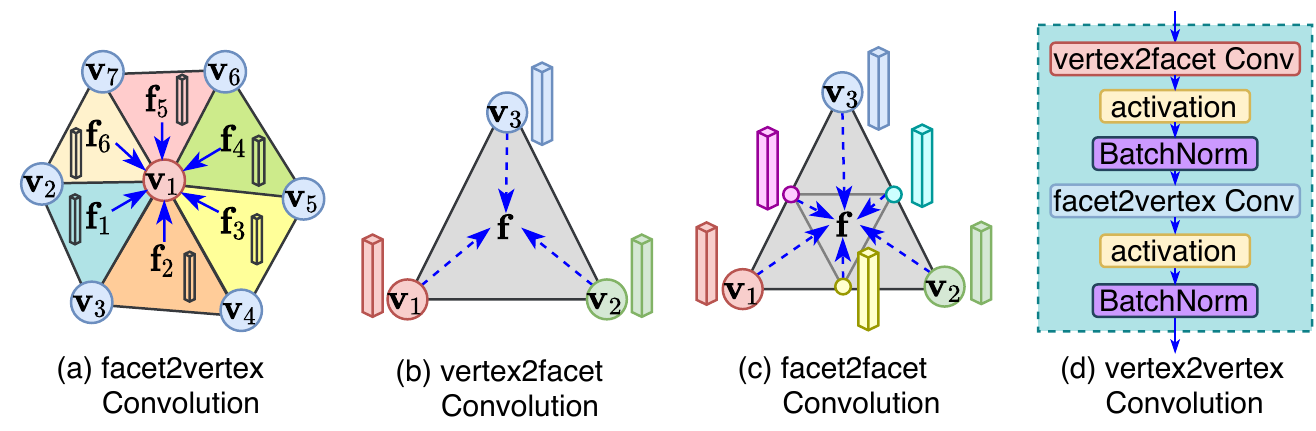}
    \vspace{-3mm}
    \caption{Mesh convolutions introduced in Picasso. 
    (a) The facet2vertex convolution propagates features from the adjacent facets of a vertex to the vertex itself. (b) The vertex2facet convolution computes the features of a facet based on its three vertices. (c) The facet2facet convolution calculates features of a rendered facet based on the vertices and interpolated points in the facet. For simplicity, we show only three interpolated points on the rendered facet. It corresponds to a setting of $\gamma=1$ and $K=6$ following Eq.~(\ref{equ:interpolate_num}). (d) The vertex2vertex convolution is composed of a vertex2facet convolution followed by a facet2vertex convolution. We apply batch normalization to both vertex and facet features. 
    } 
    \label{fig:v2v_conv}
    \vspace{-2mm}
\end{figure}

\subsection{Geometric Features}\label{subsec:mesh_facet_geometry}
Let us denote the coordinates of facet vertices by ${\bf x}_1, {\bf x}_2, {\bf x}_3$, the edge lengths of the facet as ${\boldsymbol \ell}=(\ell_1,\ell_2,\ell_3)$ and the facet normals as ${\bf n}$. We compute inner angles of a facet  ${\boldsymbol \theta}=(\theta_1,\theta_2,\theta_3)$ as 
\begin{equation}\label{equ:inner_angles}
\begin{aligned}
\theta_1 &= \frac{\langle{\bf x}_2-{\bf x}_1, {\bf x}_3-{\bf x}_1\rangle}{\ell_1\ell_3}, \\
\theta_2 &= \frac{\langle{\bf x}_1-{\bf x}_2, {\bf x}_3-{\bf x}_2\rangle}{\ell_1\ell_2}, \\
\theta_3 &= \frac{\langle{\bf x}_1-{\bf x}_3, {\bf x}_2-{\bf x}_3\rangle}{\ell_2\ell_3}.
\end{aligned}
\end{equation}
We form the input feature representation of mesh geometry as $[{\boldsymbol \ell}, {\boldsymbol \theta}, {\bf n}]$ on each facet for shapes. For real-world surface data with aligned gravitational-axis (\eg,~the $z$-axis), we form their facet geometrics as $[{\boldsymbol \ell}, {\boldsymbol \theta}, {\bf n}, {\bf h}]$, where ${\bf h}$ concatenates the heights of the three vertices, \eg,~${\bf h}=[z_1,z_2,z_3]$. 
We note that the standard input features to our network in \S~\ref{sec:network_Pi2} include both mesh geometry and textures, if textures
are available.

\section{Picasso overview}\label{sec:overview}  
\begin{figure}[t]
    \centering \includegraphics[width=0.45\textwidth]{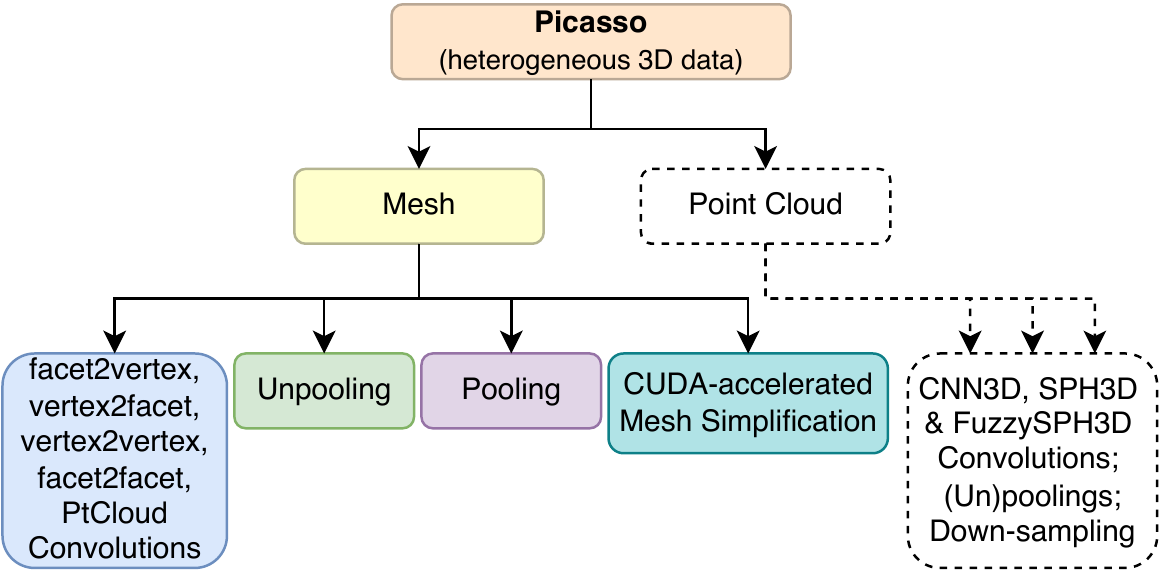}
    \vspace{-3mm}
\caption{Overview of the major deep learning modules in Picasso. We only colorize the novel modules proposed in this work. Picasso allows feature learning for both heterogeneous 3D mesh and heterogeneous 3D point cloud.}
\label{fig:picasso_overview}
\vspace{-2mm}
\end{figure}

We combine the operations proposed in this work for mesh processing  with our previously proposed operations for 3D point cloud processing in \cite{lei2020spherical,lei2020seggcn} into 
\textit{Picasso}. 
The previous point cloud operations are improved to handle point clouds of heterogeneous sizes in addition to homogeneous arrays. This improvement has led to seamless integration of our mesh and point cloud operations. Through Picasso, we make geometric deep learning over 3D data accessible to the broader research community. 
We allow easy integration of the contributed modular operations in 3D domain with the modern deep learning blocks/layers such as ResNet \cite{he2016deep}, DenseNet \cite{huang2017densely}, Inception \cite{Szegedy2015googLeNet} etc.  
Figure~\ref{fig:picasso_overview} provides an overview of the major  modules in Picasso.
To differentiate this article's contribution from  \cite{lei2020spherical,lei2020seggcn}, the figure colorizes only the novel operations introduced in this work. These include CUDA-accelerated mesh decimation, pooling, unpooling, and different mesh convolutions.
We additionally incorporate a module for GPU-based voxelization of point clouds and meshes in Picasso. The module allows mesh decimation with voxelized vertex clustering to be performed on-the-fly. 
Picasso is supported in both Pytorch \cite{paszke2019pytorch} and Tensorflow \cite{abadi2016tensorflow} for different user preferences. 
We release the code at  \href{https://github.com/EnyaHermite/Picasso}{https://github.com/EnyaHermite/Picasso}.

\vspace{1.0mm}
\subsection{\color{black}Example Usage and Heterogeneous Batching}
To build a deep convolutional block for feature learning in Picasso, 
multiple vertex2vertex convolutions can be cascaded within a network layer of the same mesh resolution, similar to the usage of CNN kernels.
In Fig.~\ref{fig:picasso_demo_usage}(left),  an example is shown for constructing a simple hierarchical mesh network using mesh convolutions and poolings.
The example network is sequentially composed of an initial convolutional layer, a max pooling layer, one convolutional block, a global pooling and an arbitrary classifier. 
Assume the network uses batch size 3 for training. Let  $\mathcal{T}_1{=}(\mathcal{V}_1,\mathcal{F}_1)$, $\mathcal{T}_2{=}(\mathcal{V}_2,\mathcal{F}_2)$,  $\mathcal{T}_3{=}(\mathcal{V}_3,\mathcal{F}_3)$ be different shapes in a batch, and $\mathcal{H}^0_1,\mathcal{H}^0_2,\mathcal{H}^0_3$ be the input features of $\mathcal{T}_1, \mathcal{T}_2, \mathcal{T}_3$, respectively. The standard input features $\mathcal{H}^0$ of a mesh comprise geometrics $\mathcal{H}_G^0$ and textures $\mathcal{H}_C^0$ on each facet. As shape meshes provided in the example do not contain textures, their input features are simplified to   $\mathcal{H}^0{=}\mathcal{H}_G^0$. In Picasso, we customize the network to accept multiple meshes via concatenation. Therefore, the shapes in the batch input are represented as a tuple of $({\bf V}, {\bf F}, {\bf H}^0)$, where
\begin{equation}
{\bf V}{=\hspace{-1mm}}
\begin{bmatrix}
\mathcal{V}_1\vspace{.5mm}\\
\mathcal{V}_2\vspace{.5mm}\\
\mathcal{V}_3\vspace{.5mm}\\
\end{bmatrix}{\hspace{-.8mm},\hspace{1.5mm}}
{\bf F}{=\hspace{-1mm}}
\begin{bmatrix}
\mathcal{F}_1{+}0\phantom{+|\mathcal{V}_1|+}\vspace{.5mm}\\
\mathcal{F}_2{+}|\mathcal{V}_1|\phantom{+|\mathcal{V}_2|}\vspace{.5mm}\\
\mathcal{F}_3{+}|\mathcal{V}_1|{+}|\mathcal{V}_2|\\
\end{bmatrix}{\hspace{-.8mm},\hspace{1.5mm}}
{\bf H}^0{=}{\bf H}_G^0{=\hspace{-1mm}}
\begin{bmatrix}
\mathcal{H}_{G,1}^0\vspace{.5mm}\\
\mathcal{H}_{G,2}^0\vspace{.5mm}\\
\mathcal{H}_{G,3}^0\\
\end{bmatrix}{\hspace{-.8mm}.}
\end{equation}

For facet concatenations in ${\bf F}$, we follow the 0-indexing convention. 
Without textures, the initial layer of the shape example consists of a $1{\times}1$ convolution followed by a facet2vertex convolution. However, the standard initial layer takes both geometrics ${\bf H}_G^0$ and textures ${\bf H}_C^0$ as input features. We show its configurations in  Fig.~\ref{fig:picasso_demo_usage}(right).
To pool the features, the mesh has to be  decimated first such that the pooling operation can proceed. We exploit max pooling in the example, while the convolutional block comprises two vertex2vertex convolutions. 
Global pooling induces a single representation for each sample such that the final classification can be applied. 
We present a basic example network in Fig.~\ref{fig:picasso_demo_usage} to provide a clear overview of Picasso. Next, we introduce our proposed network, which offers a more advanced example of using the Picasso modules.

\begin{figure*}[!t]
\centering
\includegraphics[width=0.9\textwidth]{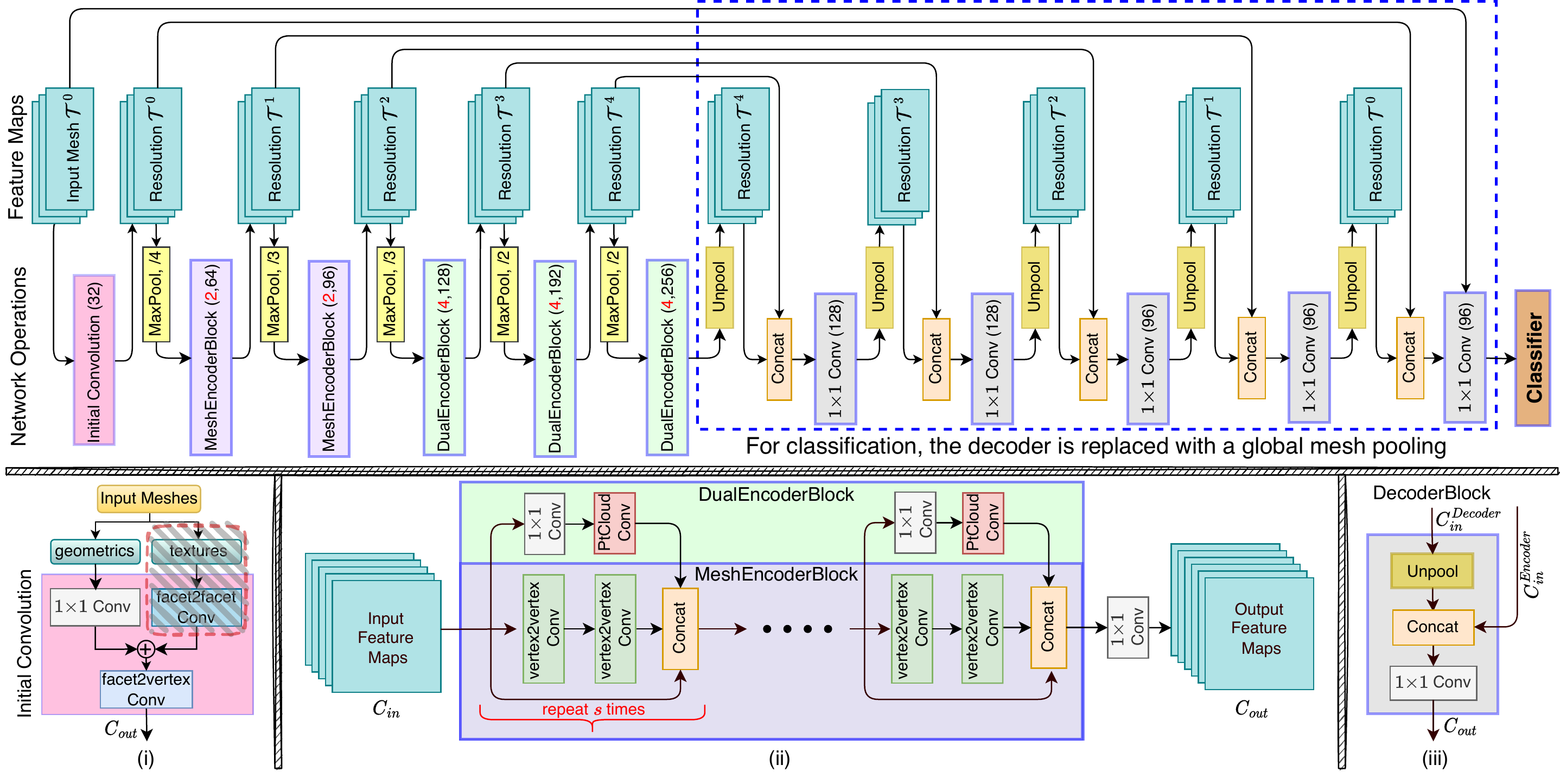}
\vspace{-3mm}
\caption{PicassoNet++ for large-scale semantic parsing of complete scenes (\textbf{top}), and its convolution blocks (\textbf{bottom}). The network consists of six mesh resolutions including the input $\mathcal{T}^{0\sim5}$. 
The output channels are respectively $32,64,96,128,192,256$ in the encoder and $128,128,96,96,96$ in the decoder. 
The pooling strides are $4,3,3,2,2$, which can be different for shape analysis. The figure depicts predicting semantics of mesh vertices,  a vertex2facet convolution can be inserted before the final classifier for facet-based predictions. For classification, the decoder is replaced by a global mesh pooling. 
The bottom row shows (i) the `Initial Convolution' 
which propagates input features from facet to vertex; (ii) the `MeshEncoderBlock' that is exploited in high-resolution layers for feature extraction, along the `DualEncoderBlock' that applies to low-resolution layers such that feature extraction can go beyond disconnected components of the mesh, and (iii) the `DecoderBlock' for feature unsampling from low-resolution meshes to high-resolution meshes. 
In the `Initial Convolution', the facet2facet convolution is not applicable if textures are not provided. We apply `MeshEncoderBlock' to   $\mathcal{T}^1,\mathcal{T}^2$ and `DualEncoderBlock' to   $\mathcal{T}^3,\mathcal{T}^4,\mathcal{T}^5$. They repeat $s{=}2$ and $s{=}4$ times, respectively. 
}
\label{fig:Pi2}
\vspace{-2mm}
\end{figure*}

\vspace{-2mm}
\section{PicassoNet++}\label{sec:network_Pi2}
Besides extending Picasso beyond the preliminary work in \cite{lei2021picasso}, this article also considerably enhances PicassoNet~\cite{lei2021picasso} to introduce a more effective network  PicassoNet++. 
Compared to \cite{lei2021picasso}, PicassoNet++ is deeper  yet faster for geometric feature learning over 3D meshes. 
We show its configuration in the Fig.~\ref{fig:Pi2}(top), which also  includes a decoder part (boxed) for dense
parsing. For classification, the decoder is replaced with an average global pooling layer. PicassoNet++ takes \textit{intact} meshes rather than mesh crops as input samples.  
We apply strided mesh decimation by specifying the expected vertex size using a stride parameter. This removes the 
constraint of fixing vertex sizes across different samples. 

{\color{black}
For decimated meshes of low resolution, 
PicassoNet++ exploits the point cloud convolution of \S~\ref{subsec:pcloud_conv} to extract features across disconnected components of the mesh, similar to \cite{lei2021picasso}.
The Euclidean neighborhood in point cloud convolution allows larger context to be established such that feature learning can go beyond geodesic connections. 
Whereas the previous methods \cite{lei2021picasso,schult2020dualconvmesh} explore Euclidean neighborhood extensively in every 
network layer, PicassoNet++ employs it only when the input mesh is significantly decimated. This results in considerable computational advantage. 
We show in \S~\ref{subsec:dual_layers_ablation} that point cloud convolution is unnecessary at high-resolution layers since the meshes are already well-connected.

Our network exploits two types of encoder blocks to extract features from the meshes of different resolutions. One is \textit{mesh} encoder block, which comprises a repetitive building unit that uses only mesh convolutions. The other is \textit{dual} encoder block, whose repetitive building unit is mesh convolutions accompanied by one point cloud convolution. We use identical feature channels for mesh and point cloud convolutions in the dual encoder blocks. PicassoNet++ employs mesh encoder blocks in high-resolution layers and dual encoder blocks in the low-resolution layers.}
Our network inherits the primary building units of PicassoNet. However, it applies skip connections to every building unit of the encoder block. Besides, it leverages the concatenation-style skip connection of DenseNet~\cite{huang2017densely}, instead of the addition-style skip connection of ResNet~\cite{he2016deep}.
Figure~\ref{fig:Pi2}(bottom) depicts the major blocks of PicassoNet++, including its initial convolution, mesh and dual encoder blocks, as well as the decoder blocks. We employ max mesh pooling to down-sample the network features.

{\color{black}
PicassoNet++ takes geometry and textures of each facet as input features. This differs from PicassoNet~\cite{lei2021picasso}, which follows point cloud networks and expects vertex features as inputs.
To propagate the input features from facet to vertex, we build the initial layer of PicassoNet++ using a $1{\times}1$ convolution with a parallel facet2facet convolution, followed by a feature fusion of addition `$\textcircled{+}$' and a facet2vertex convolution. 
For the decoder, we use  $1{\times}1$ convolutions and mesh unpooling of PicassoNet~\cite{lei2021picasso} to upsample the features. PicassoNet++ applies batch normalization to all of its modular convolutions. 
The proposed network repeats the building unit of its dual encoder blocks $s{=}$4 times, resulting in a deeper network than PicassoNet. For classification, we replace the single $1{\times}1$ convolution for the final predictions in dense parsing with two fully connected (FC) layers. 
}

\section{Experiments}\label{sec:experiment}
We establish the effectiveness of PicassoNet++ by evaluating it for (i)~shape analysis with synthetic meshes and (ii)~semantic scene surface parsing using real-world meshes. We use the 
ShapeNetCore \cite{chang2015shapenet} dataset, along with SHREC \cite{lian2011shape}, CUBE \cite{hanocka2019meshcnn}, COSEG \cite{wang2012active}, HUMAM \cite{maron2017convolutional} and FAUST \cite{bogo2014faust} 
for shape analysis. 
For real-world scene surface parsing, we employ the large-scale datasets S3DIS \cite{armeni20163d} and ScanNet \cite{dai2017scannet}. 
Each dataset is discussed with its related experiments. 
For all shape analysis experiments in \S~\ref{sec:synthetic_data}, we normalize the scales of all shape meshes, but keep the original scales of real-world scene surfaces in \S~\ref{sec:realistic_data}. 
This is because scale information is more important for real-world scenes. 
We employ the proposed facet geometrics $[{\boldsymbol \ell}, {\boldsymbol \theta}, {\bf n}]$ as default  input features for synthetic shapes, and $[{\boldsymbol \ell}, {\boldsymbol \theta}, {\bf n}, {\bf h}]$ as default geometric features for the real-world data. Additionally, real-world data also provides facet textures for experiments.

\vspace{1mm}
\noindent{\textbf{Data Augmentation:}} We apply standard geometric transformations, \eg random flipping, scaling and shifting to the mesh vertices. We perform random rotations along the gravitational axis for aligned data in ShapeNetCore, ScanNet and S3DIS, and free data rotation  along all axes for other datasets. We also randomly drop the vertices and facets of meshes to obtain more training data. 
When textures are available, we apply color shifting, jittering,
and standard photometric augmentation techniques as in image processing \cite{liu2016ssd,choy20194d}.

\vspace{1mm}
\noindent{\textbf{Network Configuration:}}
PicassoNet++ contains 6 hierarchical layers of mesh resolutions from $\mathcal{T}^0$ to $\mathcal{T}^5$ that use mesh decimation. 
We set different decimation strides for different datasets, discussed in the respective experiments.
Our network uses dual encoder blocks only at coarse resolutions $\mathcal{T}^3,\mathcal{T}^4,\mathcal{T}^5$.
The range search radii for point-based convolutions of $\mathcal{T}^3$ to $\mathcal{T}^5$ are $0.2$, $0.4$, $0.8$, respectively. {\color{black}We use spherical harmonics of degree $L{=}3$, which results in a kernel size of $T{=}16$ for all mesh convolutions and $T{=}17$ for the extended point cloud convolution. 
We train the network with Adam Optimizer \cite{kingma2015adam} and exponential decay. The initial learning rate is 0.001, with 0.98 decay rate after each epoch.}

\begin{table*}[!t]
\centering
\caption{\color{black}Shape analysis performance of our network on the synthetic datasets. 
}
\label{tab:shape_analysis}
\begin{adjustbox}{width=0.9\textwidth}
{
\begin{tabular}{l|c|c|c|c|c|c|c|c|c}
\hline
\multirow{3}{*}{Method}& \multicolumn{4}{c|}{Classification} & \multicolumn{4}{c|}{Semantic Labelling} & Correspondence \\
\cline{2-4}\cline{5-10} 
& \multirow{2}{*}{ShapeNetCore}& \multicolumn{2}{c|}{SHREC} & \multirow{2}{*}{CUBE} &  \multicolumn{3}{c|}{COSEG} & \multirow{2}{*}{HUMAN} &  \multirow{2}{*}{FAUST} \\
 \cline{3-4}\cline{6-8} 
 & & Split 16 & Split 10 &&  aliens & chairs & vases &  & \\
 \hline
 \# Train/test samples &40866/10261 &480/120 & 300/300& 5100/660& 169/29&337/60 & 252/45& 381/18& 80/20\\
 \hline
GI \cite{sinha2016deep} & --& 96.6& 88.6& -- &  --& --& --&  --&--\\
GWCNN \cite{ezuz2017gwcnn} & --& 96.6& 90.3& -- &  --& --& --&  --&--\\
PointNet++ \cite{qi2017pointnetplusplus} & --& --& --& 64.3 &   --& --& --& --&--\\
 MeshCNN \cite{hanocka2019meshcnn} & --& 98.6& 91.0& 92.2 &  96.3& 93.0& 92.4& 85.4&--\\
 PD-MeshNet \cite{milano2020primal} & --& 99.7& 99.1& 94.4&  98.2& 97.2& 95.4 & 85.6&--\\
 {\color{black}HodgeNet~\cite{smirnov2021hodgenet}} & -- & 99.2& 94.7& --&  96.0& 95.7& 90.3 & 85.0&--\\ 
 {\color{black}SubdivNet~\cite{hu2022subdivision}} &-- & \textbf{100.0}& \textbf{100.0}& \textbf{100.0}&  --& --& -- & \textbf{91.7}&--\\ 
 GCNN \cite{masci2015geodesic}& --& --& -- & --& --& --& --&  --& 65.4\\
ACNN \cite{boscaini2016learning}& --& --&-- & --& --& --& --&  --& 63.0 \\
MoNet \cite{monti2017geometric} & --& --& -- & --& --& --& --&  --& 90.0\\
PointContrast \cite{monti2017geometric} & 85.1&--& --& -- & --& --& --& --&  --\\
\hline
{\color{black}PicassoNet++ (Prop.)} &\textbf{87.3}&\textbf{100.0} & \textbf{100.0}& \textbf{100.0} &
\textbf{98.8} & \textbf{99.5} & \textbf{95.6} & 91.5 &
\textbf{100.0}\\
\hline
\multicolumn{10}{l}{$^{\dagger}$SubdivNet~\cite{hu2022subdivision} benefits from remeshing and finer geometric details of input shapes, whereas other methods do not.}
\end{tabular}
}
\end{adjustbox}
\end{table*}

\vspace{-2mm}
\subsection{Shape Analysis}
\label{sec:synthetic_data}
We evaluate our network performance on shape classification and facet labelling tasks using synthetic data. The input meshes are decimated with strides 1, 3, 2, 2, 2, respectively on ShapeNetCore and FAUST, while 1, 1.5, 1.5, 1.5, 1.5 on the other datasets due to their limited number of input vertices. 
Here, the first stride 1 indicates the $\mathcal{T}^0$ is not decimated and it is identical to $\mathcal{T}^1$. Therefore, the first pooling operation in PicassoNet++ is not applied.
We train the network using batch size 6 for FAUST, and 64 for others. A weight decay of $10^{-5}$ is applied to all datasets other than ShapeNetCore for their limited training samples.

\vspace{1mm}
\noindent\textit{1) Classification}
\vspace{1mm}\\
\noindent\textbf{ShapeNetCore:} The ShapeNetCore dataset \cite{chang2015shapenet} is a large-scale and information-rich repository of 3D models collected from online  resources. It contains around 51,000
shapes of 55 common objects. We follow the original standard split to evaluate the performance of PicassoNet++ for shape classification. In specific, the split specifies $80\%$ samples for training and $20\%$ for testing. Table \ref{tab:shape_analysis} shows that our network outperform the sparse Residual network of PointContrast \cite{xie2020pointcontrast} by 2.5\%.
This indicates the desirability of processing mesh data with  PicassoNet++ for shape analysis. We prepare the input meshes to our network by uniformly sampling 3,000 points on the raw mesh, and triangulating them using the algorithm provided by \cite{pointcloud2mesh}. 
We note that these meshes are not ideal and actual watertight meshes should result in even better performance of our network.

\vspace{1mm}
\noindent\textbf{SHREC:} The SHREC dataset \cite{lian2011shape, hanocka2019meshcnn} contains 600 watertight meshes from 30 classes, with 20 samples in each class. Shape classification is defined on split 16 and 10 of the dataset. The split number here indicates the number of training samples  per class. Following the setup in \cite{hanocka2019meshcnn}, we report the average results over three randomly generated sets. Table \ref{tab:shape_analysis} shows excellent performance of PicassoNet++.

\vspace{1mm}
\noindent\textbf{CUBE:} The CUBE
Engraving dataset \cite{hanocka2019meshcnn} includes 22 object categories with 200 mesh samples per class. Those samples are created by insetting the MPEG-7 binary
shapes \cite{latecki2000shape} into random locations of a cube. 
Each cube consists of about 250 vertices and 500 facets. Table~\ref{tab:shape_analysis} shows that our network achieves 100\% accuracy on this dataset.

\vspace{1mm}
\noindent\textit{2) Semantic Labelling}
\vspace{1mm}\\
\vspace{1mm}
\noindent\textbf{COSEG:} The COSEG dataset~\cite{wang2012active} defines semantic labelling tasks over three independent categories, \ie \textit{aliens}, \textit{chairs} and \textit{vases}.
The alien category contains 169 training samples, 29 test samples and 4 part labels. The chair category contains 337 training samples, 60 test samples and 3 part labels. The vase category contains 252 training samples, 45 test samples and 4 part labels. 
We follow \cite{milano2020primal} and evaluate our network under semantic facet labelling. 
Table \ref{tab:shape_analysis} reports the consistent superior performance of PicassoNet++.

\vspace{1mm}
\noindent\textbf{HUMAN:}
The HUMAN dataset~\cite{maron2017convolutional} defines semantic facet labelling as segmenting the human body into 8 parts, which include  \textit{head}, \textit{hand}, \textit{forearm}, \textit{upperarm}, \textit{body}, \textit{thigh}, \textit{leg} and \textit{foot}. 
It contains 381 training samples and 18 test samples. Each mesh sample is composed of 750 vertices and 1,500 facets. Table \ref{tab:shape_analysis} suggests that the segmentation result of PicassoNet++ outperforms the previous methods by a large margin. 

\vspace{1mm}
\noindent\textit{3) 3D Manifold Correspondence}
\vspace{1mm}\\
\noindent\textbf{FAUST:} The FAUST
dataset~\cite{bogo2014faust} is widely used for correspondence matching of 3D manifold meshes \cite{masci2015geodesic,boscaini2016learning,monti2017geometric}.
It consists of 10 different subjects with 10 different poses each, resulting in 100 watertight meshes with exact ground-truth correspondence. Each
shape is represented as a mesh with 6,890 vertices and 13,776 facets. The convention is to utilize the first pose of the
first subject (i.e.~the zeroth scan `000') as the reference, the first 80 shapes for training and the rest 20 shapes for testing. We follow MoNet~\cite{monti2017geometric} and formulate the correspondence task as a multi-class labelling problem. Similar to its configurations in semantic labelling, the proposed network accomplishes this correspondence labelling with softmax function. In specific, the number of classes is defined as 6,890, \ie 
the number of vertices in the reference mesh. We report the matching accuracy of different methods for correspondences without geodesic error in Table \ref{tab:shape_analysis}. 
Feature representation of our network achieves 100\% accuracy, which is considerably better than other methods.

\vspace{-2mm}
\subsection{Real-world Datasets}\label{sec:realistic_data}
Real-world scene surfaces have heterogeneous vertex and facet sizes, and  varying scales. We decimate the input meshes using strides $4$, $3$, $3$, $2$, $2$, respectively, 
to construct network layers of mesh resolutions from $\mathcal{T}^1$ to $\mathcal{T}^5$.
The network is trained with batch size 16.

\begin{table*}[!t]
\caption{\color{black}Performance of PicassoNet++ on the fifth fold (Area 5) of S3DIS dataset. It outperforms PointTransformer while using only half the number of training parameters (2.5M vs. 4.9M). Besides, our results are obtained by taking each \textit{complete} scene as input. 
}\label{tab:s3dis_seg_review} 
\vspace{-1.5mm}
\begin{adjustbox}{width=1\textwidth}
{\Huge\begin{tabular}{l|ccc|ccccccccccccc}
\hline
Method& OA& mAcc & mIoU & ceiling & floor & wall & beam & column & window & door & table & chair & sofa & bookcase & board & clutter \\
\hline
 PointNet \cite{qi2017pointnet}& - &49.0 &41.1 &88.8 &97.3 &69.8 &\textbf{0.1} &3.9 &46.3 &10.8 &58.9 &52.6  &5.9 &40.3  &26.4 &33.2\\
SEGCloud \cite{tchapmi2017segcloud} & - &57.4 &48.9 &90.1 &96.1 &69.9 &0.0 &18.4 &38.4 &23.1 &70.4 &75.9 &40.9 &58.4 &13.0 &41.6\\
Tangent-Conv \cite{tatarchenko2018tangent}& 82.5 &62.2 &52.8 &- &- &- &- &- &- &- &- &- &- &- &- &-\\
SPG \cite{landrieu2017large} & 86.4 &66.5 &58.0 &89.4 &96.9 &78.1 &0.0 &42.8 &48.9 &61.6&75.4 &84.7 &52.6 &69.8  &2.1 &52.2\\
PointCNN \cite{li2018pointcnn}& 85.9& 63.9& 57.3& 92.3& 98.2 &79.4& 0.0 &17.6& 22.8& 62.1& 74.4& 80.6& 31.7& 66.7& 62.1& 56.7\\
SSP+SPG \cite{landrieu2019point}& 87.9 & 68.2 &61.7&- &- &- &- &- &- &- &- &- &- &- &- &-\\
GACNet \cite{wang2019attention}& 87.8 & - &62.9&92.3 &98.3 &81.9 &0.0 &20.4 &59.1 &40.9 &78.5 &85.8 &61.7 &70.8 &74.7 &52.8\\
SPH3D-GCN \cite{lei2020spherical}& 87.7 &65.9 &59.5 &93.3 &97.1 &81.1 &0.0 &33.2 &45.8 &43.8  &79.7 &86.9 &33.2 &71.5  &54.1 &53.7\\
SegGCN \cite{lei2020seggcn} & 88.2 &70.4 &63.6 &93.7 &\textbf{98.6} &80.6 &0.0
&28.5 &42.6 &74.5  &80.9
&88.7
&69.0
&71.3  &44.4 &54.3\\
 MinkowskiNet \cite{choy20194d} & -& 71.7&65.3 &- &- &- &- &- &- &- &- &- &- &- &- &-\\
KPConv \cite{thomas2019kpconv}& - & 72.8 &67.1 &92.8& 97.3& 82.4 &0.0& 23.9& 58.0& 69.0& 81.5& \textbf{91.0}& \textbf{75.4}& 75.3 & 66.7& 58.9\\
DCM-Net \cite{schult2020dualconvmesh} & - & 71.2 & 64.0 &92.1 &96.8 &78.6 &0.0 &21.6 &61.7 &54.6 &78.9 &88.7 &68.1 &72.3 &66.5 &52.4\\
{\color{black}PCT~\cite{guo2021pct}} &-&67.7 & 61.3 & 92.5 &98.4 &80.6 &0.0 & 19.4 & 61.6& 48.0& 85.2& 76.6& 67.7& 46.2&  67.9& 52.3 \\
PointTransformer\cite{zhao2021point} & 90.8& 76.5& 70.4& 94.0& 98.5& 86.3& 0.0& 38.0& 63.4& 74.3& \textbf{89.1}& 82.4& 74.3& \textbf{80.2}& 76.0& 59.3\\
\hline
{\color{black}PicassoNet~\cite{lei2021picasso}} & 89.4 &70.9 &64.6 &93.3 &97.7 &83.5 &0.0
&31.9 &53.4 &69.2  &81.7
&88.0
&50.5
&74.3  &58.2 &57.9\\
{\color{black}PicassoNet++ (Prop.)} & \textbf{91.3} &\textbf{77.2} &\textbf{71.0} &\textbf{94.4}&98.4 &\textbf{87.5} &0.0 &\textbf{46.9} &\textbf{63.7} &\textbf{75.5}  &81.4 &90.3
&71.3&76.2 &\textbf{76.7}&\textbf{61.1}\\
\hline
\end{tabular}}
\end{adjustbox}
\end{table*}

\begin{table*}[!t]
\centering
\caption{\color{black}Semantic vertex labelling results on the test set of ScanNet. The training parameters of KPConv, MinkowskiNet, and DCM-Net are 25.6M, 29.8M, 76.1M, respectively, whereas PicassoNet++ uses only 2.5M parameters. }
\label{tab:scannet_test}
\vspace{-1.5mm}
\begin{adjustbox}{width=1\textwidth}
{\Huge\begin{tabular}{l|c|cccccccccccccccccccc}
\hline
Method & mIoU & floor &wall &chair &sofa &table& door& cab& bed &desk &toil &sink &wind& pic &bkshf &curt &show &cntr &fridg& bath &other\\
\hline
SPLATNET$_{\text{3D}}$ \cite{su2018splatnet}& 39.3&92.7&69.9&65.6&51.0&38.3&19.7&31.1&51.1&32.8&59.3&27.1&26.7&0.0&60.6&40.5&24.9&24.5&0.1&47.2&22.7\\
Tangent-Conv~\cite{tatarchenko2018tangent}& 43.8&91.8&63.3&64.5&56.2&42.7&27.9&36.9&64.6&28.2&61.9&48.7&35.2&14.7&47.4&25.8&29.4&35.3&28.3&43.7&29.8\\
PointCNN~\cite{li2018pointcnn} &45.8&94.4&70.9&71.5&54.5&45.6&31.9&32.1&61.1&32.8&75.5&48.4&47.5&16.4&35.6&37.6&22.9&29.9&21.6&57.7&28.5\\
PointConv~\cite{wu2019pointconv}& 55.6&94.4&76.2&73.9&63.9&50.5&44.5&47.2&64.0&41.8&82.7&54.0&51.5&18.5&57.4&43.3&57.5&43.0&46.4&63.6&37.2\\
SPH3D-GCN~\cite{lei2020spherical}& 61.0&93.5&77.3&79.2&70.5&54.9&50.7&53.2&77.2&57.0&85.9&60.2&53.4&4.6&48.9&64.3&70.2&40.4&51.0&85.8&41.4\\
KPConv~\cite{thomas2019kpconv}& 68.4&93.5&81.9&81.4&78.5&61.4&59.4&64.7&75.8&60.5&88.2&\textbf{69.0}&63.2&18.1&78.4&77.2&80.5&47.3&58.7&84.7&45.0 \\
SegGCN~\cite{lei2020seggcn}& 58.9&93.6&77.1&78.9&70.0&56.3&48.4&51.4&73.1&57.3&87.4&59.4&49.3&6.1&53.9&46.7&50.7&44.8&50.1&83.3&39.6\\
{\color{black}CBL~\cite{tang2022contrastive}} & 69.3 &	74.3	&79.4 &	65.5	&68.4	&82.2	&49.7	&\textbf{71.9}	&62.2	&61.7	&\textbf{97.7}	&44.7	&33.9	&75.0	&66.4	&70.3	&79.0	&\textbf{59.6}	&94.6	&85.5	&64.7 \\
MinkowskiNet \cite{choy20194d} & \textbf{73.6}&95.1&85.2&84.0&77.2&\textbf{68.3}&\textbf{64.3}&70.9&\textbf{81.8}&\textbf{66.0}&87.4&67.5&72.7&28.6&\textbf{83.2}&\textbf{85.3}&\textbf{89.3}&52.1&\textbf{73.1}&\textbf{85.9}&\textbf{54.4} \\
DCM-Net~\cite{schult2020dualconvmesh} & 65.8&94.1&80.3&81.3&72.7&56.8&52.4&61.9&70.2&49.4&82.6&67.5&63.7&\textbf{29.8}&80.6&69.3&82.1&46.8&51.0&77.8&44.9\\
\hline
{\color{black}PicassoNet++ (Prop.)} & 69.2&\textbf{95.2}&\textbf{85.4}&\textbf{86.6}&\textbf{81.0}&56.4&62.6&67.7&77.2&50.9&90.3&68.9&\textbf{72.9}&22.5&78.6&84.8&70.4&51.7&54.5&73.2&53.6 \\
\hline
\end{tabular}}
\end{adjustbox}
\end{table*}

\vspace{1mm}
\noindent\textbf{S3DIS.} The Stanford 3D Indoor Spaces (S3DIS) dataset~\cite{armeni20163d} is a large-scale real-world dataset. It has sparse 3D meshes and dense 3D point clouds of 6 large-scale indoor areas. The data was collected, using the Matterport scanner, from three different buildings in Stanford University campus. 
The semantic labelling task on this dataset is defined to classify 13  classes, namely \emph{ceiling, floor, wall, beam, column, window, door, table,
chair, sofa, bookcase, board}, and \emph{clutter}. We follow the standard training/testing protocol where Area 5 is used as the test set and the remaining 5 Areas as the training set \cite{landrieu2017large,li2018pointcnn,qi2017pointnet}. 
Performance of each method is evaluated  for Overall Accuracy (OA), mean  Accuracy of all classes (mAcc),   Intersection Over Union of each class (IoU) and their average over all classes (i.e.~mIoU). mIoU is normally considered the most reliable among these metrics. 

DCM-Net \cite{schult2020dualconvmesh} prepared its training meshes and labels based on the original meshes with over-tessellation and interpolation. In contrast, we generate the scene meshes by triangulating the labelled point cloud provided in the dataset.  In specific, we voxelize the raw point cloud using a voxel size of 0.03 (3$cm$), and triangulate them into meshes \cite{pointcloud2mesh}. We guarantee all of the created meshes to be \textit{edge-manifold}. In this experiment, we utilize the default facet geometrics together with rendered facets of texture resolutions determined by $(\alpha,\beta)=(3,1)$ as input features to PicassoNet++.
We train and test the network using complete scenes as input samples. It can be noticed from Table~\ref{tab:s3dis_seg_review} that 
our method significantly outperforms the previous methods. 
{\color{black}The average inference time of PicassoNet++ in Pytorch is 170 ms
across the 68 (voxelized) test samples in Area~5, using a single NVIDIA 3090 GPU.} The final results reported in Table~\ref{tab:s3dis_seg_review} are computed on the original point cloud. We transfer the voxelized predictions to dense predictions using nearest neighborhood search.

\vspace{2mm}
\noindent\textbf{ScanNet.}  The ScanNet dataset~\cite{dai2017scannet} comprises 
reconstructed room meshes from RGB-D video frames, and has rich annotations for semantic vertex labelling.
It includes 1,613 meshes in total, among which 1,213 scenes are used for training and 300 scenes for validation. We ignore the 100 test samples in our experiment as their labels are unavailable. The dataset contains 40 class labels, while 20 are recommended for performance evaluation. We train and test our network without cropping the complete scenes into smaller samples.

Our network takes voxelized mesh of grid size 2$cm$ as the input. 
{\color{black}Similar to the experiments on S3DIS, we use the geometric features together with rendered textures of each facet as input features. The texture resolution is  $(\alpha,\beta)=(3,3)$.
Our results on the validation set of ScanNet is 71.8\%,} which is 3.6\% higher than DCM-Net and is very competitive to the 72.2\% of the top performer MinkowskiNet. MinkowskiNet has 29.8M training parameters \cite{xie2020pointcontrast}, while our network produces similar results using just {\color{black}2.5M parameters}. We report the results of our network on the test benchmark of ScanNet in Table~\ref{tab:scannet_test}, which validates that PicassoNet++ is very competitive to the top performer. {\color{black}We note that 
PicassoNet++ takes 180 ms on average in Pytorch to process per (voxelized) mesh on a single NVIDIA RTX 3090 GPU.}

\section{Further Analysis}
\label{sec:ablation}
In this section, we provide further results to analyze the proposed approach. 
\vspace{-2mm}
{\color{black}
\subsection{Varying the Degree of Spherical Harmonics}
We study influence of the degree $L$ of the spherical harmonics on the performance of PicassoNet++. The HUMAN dataset is utilized in this analysis. Table~\ref{tab:diff_spharm_degree} summarizes the resulting network parameters and segmentation accuracy for different values of $L$. It can be noticed that generally, larger values of $L$ (\eg,~$2$, $3$, $4$) result in more accurate predictions because they introduce more basis functions and parameters, leading to a better fit of a continuous filter. However, too large values of $L$ (\eg,~$5$) can cause overfitting. By default, PicassoNet++ uses $L=3$. 

\begin{table}[t]
\centering
\caption{\color{black}Performance of PicassoNet++ on the \\HUMAN dataset using different degrees \\of spherical harmonics.}\label{tab:diff_spharm_degree}
\vspace{-1.5mm}
\begin{adjustbox}{width=0.48\textwidth}
{
\large
\begin{tabular}{l|c|c|c|c|c|c}
\hline
Degree ($L$) & 5 & 4 & 3 &2 & 1 & 0\\
\hline 
\#Parameters &$2.81M$ & $2.67M$& $2.56M$& $2.48M$ & $2.42M$ & $2.38M$ \\ 
\hline
Accuracy & 91.0 & 91.4& 91.5 & 91.6& 90.9& 91.0\\
\hline
\end{tabular}
}
\end{adjustbox}
\vspace{-2mm}
\end{table}}

\vspace{-2mm}
\subsection{Is Dual Convolution Always Necessary?}\label{subsec:dual_layers_ablation}
It is known that point cloud convolutions can be time-consuming because of neighborhood search, which accompanies a significant computational burden in processing dense data~\cite{lei2020spherical}. 
The DCM-Net and original PicassoNet~\cite{lei2021picasso}  use dual convolutions in every layer of their networks. 
In comparison, PicassoNet++ utilizes dual convolutions only in its encoder blocks of coarse resolutions~$\mathcal{T}^{3\sim5}$. 
To consolidate our pruning of point-based convolutions for PicassoNet++, we empirically evaluate if dual convolution is  necessary for every layer. In specific, we alter the PicassoNet++ by either adding point cloud convolutions to its encoders of resolution~$\mathcal{T}^2$, or removing its existing point cloud convolution from the encoder of resolution~$\mathcal{T}^3$.
We test the performance of these variants, and report their results as well as other details in Table~\ref{tab:nLayer_dual}. ScanNet dataset is utilized in this experiment. From our findings, we can conclude that point-based convolutions can be eliminated from high resolution layers of the network, without affecting the network performance.  This led us to our eventual configuration of PicassoNet++, which is both effective and efficient. The reported inference time in the Table is for voxelized meshes of grid size 2$cm$. To further confirm our finding, we also conducted a similar  experiment on the HUMAN dataset. The results in Table~\ref{tab:exp_human_dual} validate  the passive role of point cloud convolutions for dense meshes.

\begin{table}[!t]
    \centering
     \caption{\color{black}Performance and runtime of PicassoNet++ on ScanNet while adding or deleting point cloud convolutions.  The list of NN search radii denotes the radii of neighborhood search for point cloud convolutions from resolution $\mathcal{T}^2$ to $\mathcal{T}^5$.}
    \label{tab:nLayer_dual}
    \vspace{-1.5mm}
    \begin{adjustbox}{width=0.47\textwidth}{
    \begin{tabular}{l|c|c|c}
    \hline
    Config & adding & \textbf{used} & reducing \\
    \hline
      Dual Levels &  $(\mathcal{T}^2, \mathcal{T}^3, \mathcal{T}^4, \mathcal{T}^5)$ & $(\mathcal{T}^3, \mathcal{T}^4, \mathcal{T}^5)$& $(\mathcal{T}^4, \mathcal{T}^5)$  \\
      \hline
      NN search radii & (0.1,~0.2,~0.4,~0.8)& (0.2,~0.4,~0.8)&  (0.4,~0.8) \\ 
      \hline
      Inference time (ms) & 240 & 180 & 160 \\
      \hline 
      mIoU  & 71.8 & 71.8 & 71.0 \\
      \hline
      \multicolumn{4}{l}
      {$^{\dagger}$Network runtime is reported in Pytorch.}
    \end{tabular}}
    \end{adjustbox}
\end{table}

\begin{table}[!t]
\centering
\caption{\color{black}PicassoNet++ performance on the HUMAN dataset while adding or deleting point cloud convolutions.
}
\label{tab:exp_human_dual}
 \vspace{-1.5mm}
\begin{adjustbox}{width=0.48\textwidth}
{
\begin{tabular}{l|c|c|c|c|c}
\hline
 Dual &  $(\mathcal{T}^2, \mathcal{T}^3, \mathcal{T}^4, \mathcal{T}^5)$ &  $(\mathcal{T}^3, \mathcal{T}^4, \mathcal{T}^5)$& $(\mathcal{T}^4, \mathcal{T}^5)$ & $(\mathcal{T}^5)$ & None\\
\hline
Radii & (0.1,~0.2,~0.4,~0.8) & (0.2,~0.4,~0.8)&  (0.4,~0.8) & (0.8) & N.A. \\ 
\hline
Acc &91.4& 91.5 & 91.1 & 90.4 & 88.7   \\
\hline
\end{tabular}
}
\end{adjustbox}
\end{table}

\subsection{2D Embedding of the Shape Features}
\begin{figure}[!t]
    \centering
\hspace{-1mm}\includegraphics[width=0.49\textwidth]{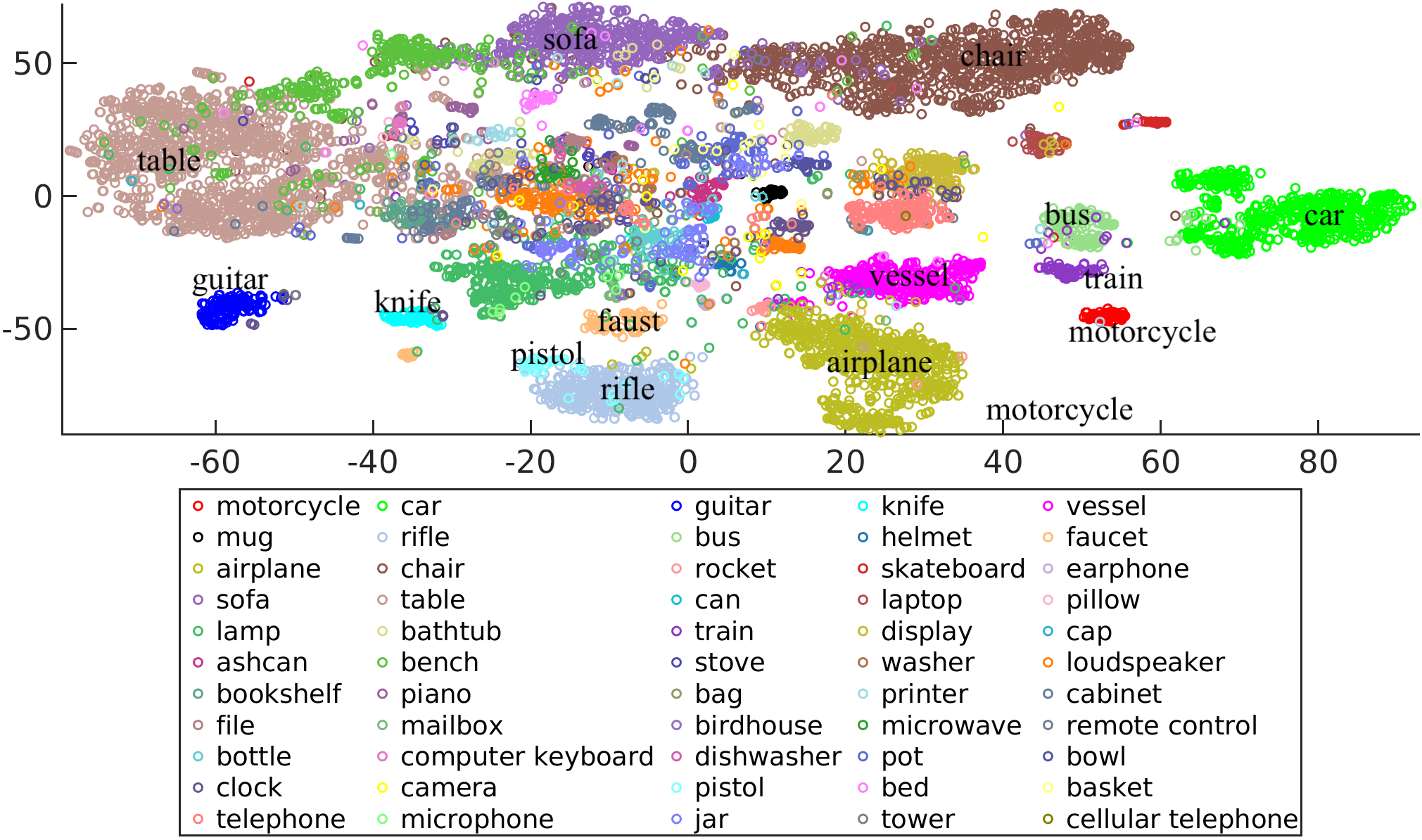} \\
    \vspace{-2mm}
    \caption{\color{black}2D embedding of the shape feature representations learned by PicassoNet++ for the test  samples of ShapeNetCore.}
    \label{fig:tSNE_2Dembed_ShapeNetCore}
\end{figure}

We visualize the shape features learned by PicassoNet++ for the test samples of ShapeNetCore \cite{chang2015shapenet} by showing their 2D embeddings in Fig.~\ref{fig:tSNE_2Dembed_ShapeNetCore} using the t-SNE technique \cite{van2008visualizing} for dimension reduction. For the feature representations, we use the 256-dimensional output of global pooling in the classification network. 
From Fig.~\ref{fig:tSNE_2Dembed_ShapeNetCore}, it is clear that the shapes of most classes are distinctly represented, such as \textit{car}, \textit{bus},  \textit{guitar}, \textit{knife}, \textit{vessel}, \textit{rifle}, \textit{faucet},  \textit{airplane}, \textit{chair}, \textit{sofa}, \textit{table}, etc. We also note that some classes are much closer to each other such that their  closeness is well-justified based on their shapes and semantics. For instance, buses are close to cars and pistols are close to rifles.

\subsection{Semantic Parsing Visualization}
\begin{figure}
    \centering
    \includegraphics[width=0.49\textwidth]{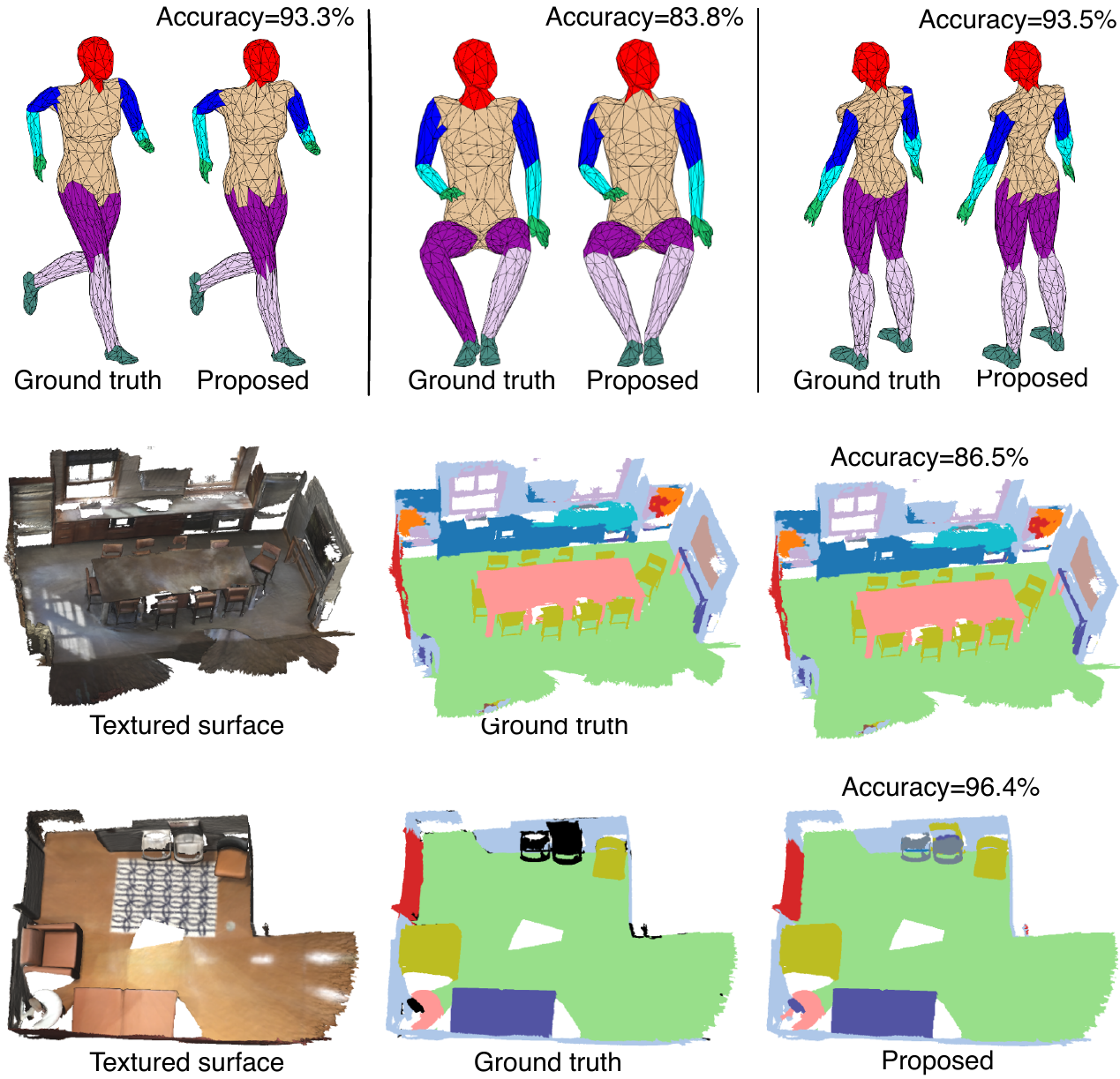}
    \vspace{-2mm}
    \caption{\color{black}The ground truth and our predictions for the shapes of human bodies and real-world scene surfaces. The black colors in the ground truth of textured surfaces indicate unlabelled objects.}
    \label{fig:sem_vis}
     \vspace{-2mm}
\end{figure}

As representative examples, we visualize the semantic parsing results of PicassoNet++ for shapes of human bodies \cite{maron2017convolutional} and surfaces of real-world scenes \cite{dai2017scannet} in Fig.~\ref{fig:sem_vis}.
The network predicts most of the body parts and scene objects correctly. 
However, we see that segmenting parts and objects near boundaries sometimes  cause minor issues for our network. Nevertheless, such errors remain minor and do not occur too frequently. 
Also notice that one of the test samples of human bodies has incorrect ground-truth label for the right leg. Such ground truth problems can result in a lower accuracy value of an accurate technique like ours. This also indicates that instead of highest prediction performance on a single dataset, highly competitive results across multiple datasets is sometimes more preferable in this domain. PicassoNet++ is able to achieve that.  

{\color{black}
\section{Limitations}
Our proposed modules and neural network focus on triangle meshes as inputs. Whereas meshes are often readily available, in some cases, \eg, outdoor settings~\cite{hackel2017semantic3d}, \cite{behley2019semantickitti}, it is not the case. 
The proposed mesh processing modules are obviously not applicable to those cases.  
Fortunately, significant progress has been made in the field of surface reconstruction utilizing deep learning techniques such as NeRF~\cite{mildenhall2020nerf,tancik2022block}, SDF~\cite{park2019deepsdf}, and deep computational geometry~\cite{chen2022neural,lei2023circnet}. This progress is likely to make meshes more accessible in the future, which will resolve this limitation. 

All of our proposed convolutions are translation-invariant. Furthermore, the vertex2facet and facet2facet convolutions are also rotation-invariant, as they utilize Barycentric coordinates. However, the facet2vertex convolution and the extended point cloud convolution do not hold rotation-invariance due to their dependence on normals and relative locations between the points. This is an intrinsic limitation of the underlying objective of this convolution, which must be considered in its application.  
}

\vspace{-3mm}
\section{Conclusion}
We made two major contributions towards hierarchical neural modeling of heterogeneous 3D meshes. First, we
presented Picasso - a modular implementation of multiple desired operations for geometric feature learning over 3D meshes. Picasso introduces novel mesh-amenable convolutional operations, mesh (un)poolings and GPU-accelerated mesh decimation. {\color{black}This article considerably enhances our preliminary version of Picasso by incorporating continuous filter modelling and improved efficiency. Moreover, we also release Pytorch version of Picasso with this article along the Tensorflow support. The second major contribution of this article is our network, PicassoNet++. Enabled by the upgraded Picasso, our network is able to effectively process mesh signals, including primitive geometrics and textures on the facets, as inputs.} It also takes advantage of a new insight provided in this article regarding the passive role of point cloud convolutions in high resolution mesh feature learning. Leveraging that, PicassoNet++ learns features over 3D shapes and scene surfaces efficiently. Through extensive experiments, we established the highly competitive performance of PicassoNet++ for shape analysis and scene parsing.

\end{document}